\documentclass[preprint]{elsarticle}
\usepackage[hidelinks,bookmarks=false]{hyperref}
\usepackage[
singlelinecheck=false
]{caption}
\usepackage[utf8]{inputenc}
\usepackage[T1]{fontenc}
\usepackage{subfigure}
\usepackage[DIV11]{typearea} %TODO nur für arxiv!
\usepackage{amsmath,amssymb,amsfonts}

\DeclareMathOperator*{\argmin}{arg\,min}
\usepackage[ruled,linesnumbered, noend]{algorithm2e}
\usepackage{wrapfig,lipsum,booktabs}
\usepackage{graphicx}
\usepackage{textcomp}
\usepackage{threeparttable}
\usepackage[]{xcolor}
\usepackage{tabularx,booktabs}
\usepackage{siunitx}
\usepackage{tikz}

 \usetikzlibrary{calc, trees, positioning, arrows, shapes, shapes.multipart, shadows, matrix, decorations.pathreplacing, decorations.pathmorphing}
\usepackage{pgfplots}
\usepgfplotslibrary{colormaps,external}
\pgfplotsset{compat=newest}
\def\layersep{2.5cm}
\def\rightpic{16cm}

\usepackage{etoolbox}
\robustify\bfseries
\newrobustcmd{\hl}{\color{gray}}

\newcommand{\R}{\mathbb{R}}

\newcommand{\decay}{\mathit{decay}}

\definecolor{unigreen}{RGB}{153,204,51}
 \definecolor{unidarkgreen}{RGB}{74,106,11}
\definecolor{unilightgreen}{RGB}{244,255,227}

\definecolor{mathblue}{RGB}{100,150,202}
\definecolor{mathdarkblue}{RGB}{0,61,119}

\def\BibTeX{{\rm B\kern-.05em{\sc i\kern-.025em b}\kern-.08em
    T\kern-.1667em\lower.7ex\hbox{E}\kern-.125emX}}

\begin{document}
\begin{frontmatter}
    \title{Efficient and Sparse Neural Networks by Pruning Weights in a Multiobjective Learning Approach}    
    \author[1]{Malena Reiners}
    %\ead{reiners@math.uni-wuppertal.de}
    \author[1]{Kathrin Klamroth}
    %\ead{klamroth@math.uni-wuppertal.de}
    \author[1]{Michael Stiglmayr}
    %\ead{stiglmayr@math.uni-wuppertal.de}
    \address[1]{School of Mathematics and Natural Sciences, University of Wuppertal, Germany \\ \vspace*{0.1cm}$\{$reiners, klamroth, stiglmayr$\}$@math.uni-wuppertal.de}
   
\begin{abstract} 
Overparameterization and overfitting are common concerns when designing and training deep neural networks, that are often counteracted by pruning and regularization strategies. However, these strategies remain secondary to most learning approaches and suffer from time and computational intensive procedures. We suggest a multiobjective perspective on the training of neural networks by treating its \emph{prediction accuracy} and the \emph{network complexity} as two individual objective functions in a biobjective optimization problem. As a showcase example, we use the cross entropy as a measure of the prediction accuracy while adopting an $l_1$-penalty function to assess the total cost (or complexity) of the network parameters. The latter is combined with an intra-training pruning approach that reinforces complexity reduction and requires only marginal extra computational cost.
From the perspective of multiobjective optimization, this is a truly large-scale optimization problem. We compare two different optimization paradigms: On the one hand, we adopt a scalarization-based approach that transforms the biobjective problem into a series of weighted-sum scalarizations. On the other hand we implement stochastic multi-gradient descent algorithms that generate a single Pareto optimal solution without requiring or using preference information. In the first case, favorable knee solutions are identified by repeated training runs with adaptively selected scalarization parameters.
Preliminary numerical results on exemplary convolutional neural networks confirm that large reductions in the complexity of neural networks with neglibile loss of accuracy are possible. 
    \end{abstract}

\end{frontmatter}

\section{Introduction}\label{sec:intro}
\subsection{Motivation}
%1. Problemstellung 
Deep neural networks (DNNs) use an immense number of parameters and therefore require powerful computer hardware during operation, and even more so for the initial training of the network. As a consequence, they become impractical when only limited hardware resources and/or time are available. Furthermore, with a large number of weights and nodes there is an increasing risk of overfitting. In addition to the neural network's parameters, many other components and hyperparameters are to be selected and fine-tuned when designing and training neural networks, e.g., the type of layers, the batch size, the amount of dropout or the regularization and the learning rate. 
This pre-training processes also account for a large part of extra costs of training neural networks, which increases with its number.
Neural network training thus has to compromise between at least two conflicting goals: On one hand, the \emph{prediction accuracy} should be as high as possible (which usually asks for many network parameters and sophisticated hyperparameter settings), while on the other hand, the \emph{network complexity} should be as low as possible and the training should require a minimum number of parameters. 
\begin{figure}[tbp]
  \centering
  \begin{tikzpicture}[shorten >=1pt,->,draw=black!75, node distance=\layersep,scale=0.4]
    \tikzstyle{every pin edge}=[<-,shorten <=1pt]
    \tikzstyle{neuron}=[circle,fill=black!25,minimum size=7pt,inner sep=0pt]
    \tikzstyle{input neuron}=[neuron, fill=mathdarkblue];
    \tikzstyle{output neuron}=[neuron, fill=mathdarkblue];
    \tikzstyle{hidden neuron}=[neuron, fill=mathblue];

    \foreach \name / \y in {1,...,4}
        \node[input neuron, pin=left:] (I-\name) at (0,-\y) {};

    \foreach \name / \y in {1,...,7}
        \path[yshift=1.5cm]
            node[hidden neuron] (H-\name) at (\layersep,-\y cm) {};
    \foreach \name / \y in {1,...,5}
        \path[yshift=0.5cm]
            node[hidden neuron] (HH-\name) at (2*\layersep,-\y cm) {};
    \node[output neuron] (O) at (3*\layersep,-2.5 cm){};

    \foreach \source in {1,...,4}
        \foreach \dest in {1,...,7}
            \path (I-\source) edge (H-\dest);
    
    \foreach \source in {1,...,7}
        \foreach \dest in {1,...,5}
            \path (H-\source) edge (HH-\dest);

    \foreach \source in {1,...,5}
        \path (HH-\source) edge (O);

\node[single arrow, rounded corners=1pt, fill=unigreen, xshift=4.3cm, yshift=-0.95cm, draw, align=center, minimum height=6, minimum width=1.5]{\scriptsize\sffamily Optimization\phantom{\raisebox{-1ex}{\rule{1pt}{3.8ex}}}};

    \foreach \name / \y in {1,...,4}
        \node[input neuron, pin=left:] (I-\name) at (\rightpic,-\y) {};

    \foreach \name / \y in {1,...,5}
        \path[yshift=0.5cm]
            node[hidden neuron] (H-\name) at (\rightpic+\layersep,-\y cm) {};

    \foreach \name / \y in {1,...,4}
        \path
            node[hidden neuron] (HH-\name) at (\rightpic+2*\layersep,-\y cm) {};

    \node[output neuron] (O) at (\rightpic+3*\layersep,-2.5) {};

    \path (I-1) edge (H-1);
%     \path  (I-1) edge (HH-1);
    \path (H-1) edge (HH-1);
    \path (I-1) edge (H-2);
   % \path (I-1) edge (H-3);
    \path (I-2) edge (H-2);
    \path (I-2) edge (H-3);
   % \path (I-2) edge (H-4);
%     \path [out=-25,in=-175] (I-3) edge (HH-4);
    \path (I-3) edge (H-4);
    %\path (I-3) edge (H-5);
%     \path [out=-35,in=-145] (I-4) edge (HH-4);
    \path (I-4) edge (H-4);
    \path (I-4) edge (H-5);
    
    \path (H-2) edge (HH-1);  
    \path (H-2) edge (HH-2);
    \path (H-2) edge (HH-3);  
    \path (H-3) edge (HH-2);
    \path (H-3) edge (HH-3);
    \path (H-4) edge (HH-2);
    \path (H-4) edge (HH-3);
    \path (H-5) edge (HH-4);
    
    \foreach \source in {1,...,4}
        \path (HH-\source) edge (O);

\end{tikzpicture}
  \caption{Towards efficient and sparse network architectures.\label{fig:effnetstruct}}
\end{figure}
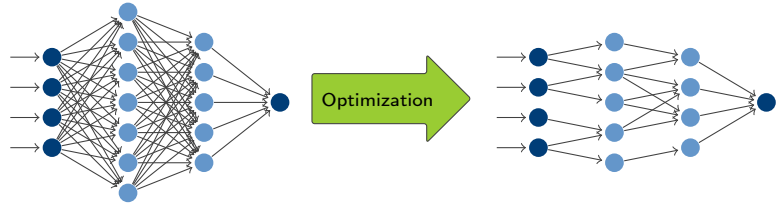 
%2.Lösungsmethode(n)
\emph{Network pruning} methods are an effective approach to limit overparameterization of DNNs and to reduce the complexity of the network. Pruning removes edges (weights), nodes (neurons) or even feature maps (filters) from the network according to their importance. Figure~\ref{fig:effnetstruct} illustrates the desired transformation from highly complex and dense networks to \emph{efficient and sparse} network designs. Removing parts of the network reduces the required storage capacity and speeds up the inferece time of the network (response time). 
Compared to initializing the search already with small networks, training and pruning of complex network structures often yields a better performance. This is reported in several recent studies, see, e.g., \citet{Li2016} and \citet{He2017}. For the success of network pruning the hyperparameter setup, including how much regularization is applied, plays a crucial role. To shorten the time-consuming search for promising hyperparameter values in general, automated recognition algorithms are suggested, e.g., by \citet{Domhan2015}. As pruning can be seen as a learning of the network architecture used for the actual learning process, it also relates to the concept of \emph{metalearning}. The following section reviews the related literature.

%4. Eigener Ansatz
In this paper, we take a \emph{biobjective} perspective on network training and consider the network complexity as an equitable objective function rather than as a regularization term. We argue that this is the basis for a thorough analysis of the \emph{trade-off} between prediction accuracy on one hand and network complexity on the other hand, and it supports the selection of preferable network architectures. We use the \emph{cross entropy} as a classical \emph{loss function} to model the prediction accuracy.
In order to minimize network complexity, we assess the relevance of individual weights in dense layers and apply pruning \emph{during} the actual training procedure. Since the relevance of individual weights has been shown to be correlated to their absolute values, see, e.g., \citet{Mummadi2019}, we use the $l_1$-norm of the weight vector as an efficiently computable second training objective. Our goal is to guide the search towards weight values that are closer to zero. At fixed intervals during the training, weights that fall below a threshold value are tentatively pruned in order to strengthen the regularizing effect of the $l_1$-objective function. 

In order to analyze the pros and cons of this multiobjective perspective on neural network training, we adopt and combine a variety of different strategies commonly used in multiobjective optimization and/or machine learning. Crucial components are (i) the implementation of the second, complexity-related objective function, and (ii) the approximation of the most promising Pareto optimal solutions of the biobjective problem. While (i) involves a sophisticated combination of complexity assessment (realized through an $l_1$-objective) and pruning (realized by an intra-training strategy), we adopt both a weighted sum approach and a variant of a stochastic multi-gradient descent algorithm for (ii). 
In the weighted sum approach we particularly analyze the influence of the constant weighting parameter. Thereby we are interested in those weightings of the two objectives for which a knee solution is found. 

The actual training is implemented based on variants of a single-objective \emph{stochastic gradient descent} (SGD) algorithm (see \citet{Robbins1951} for an early reference and \citet{Leon1999} for its application in machine learning) and of its multiobjective counterpart, a \emph{stochastic multi-gradient descent} (SMGD) algorithm suggested by \citet{Liu2019}. We compare vanilla implementations of SGD and SMGD with their respective variants including adaptive moment estimation (Adam) by \citet{kingma14adam} and root mean square propagation (RMSProp) by \citet{GHinton}, respectively.
A short review of these optimizers is given in Section~\ref{sec:optim}. 
All optimizers are compared and analyzed in relation to the chosen learning rate and other hyperparameters used. To demonstrate the efficiency of our intra-training pruning strategy, we limit ourselves to image classification problems and focus on the structural design of dense layers in \emph{convolutional neural networks} (CNNs). 

%5. Vorteile des eigenen Ansatzes
The main advantage of this novel approach can be seen in the fact that the network quality and the network architecture are optimized consistently and simultaneously. This is highly efficient and saves significant amounts of time and computing costs. Our pruning strategy does not require any fine tuning before or after the training while reaching a reduction of almost 99\% in nonzero weights, for example, in a dense layer of the CNN model for training on the image classification dataset CIFAR10 by \citet{Krizhecsky2009}. The resulting networks are sparse and can be evaluated efficiently. They are thus suitable for different types of limited hardware devices including, e.g., embedded systems. To the best of our knowledge, these are the first results using SMGD for a Keras \cite{chollet2015keras} neural network architecture which considers regularization as a second objective function. It is based on custom implementations\footnote{The python code to follow up and reproduce our results can be found at: \href{https://github.com/malena1906/Pruning-Weights-with-Biobjective-Optimization-Keras}{https://github.com/malena1906/Pruning-Weights-with-Biobjective-Optimization-Keras}} adapted from the Keras SGD, Adam and RMSProp methods and the SMGD algorithm suggested by \citet{Liu2019}. 
Furthermore, the combination of pruning with a biobjective training approach compromising between loss and regularization has not been investigated before.

\subsection{Related Research}

%1. Kurz geschichtlicher verlauf 
Many methods have been developed to deal with the problem of overparameterization in DNNs and in particular CNNs. 
Examples include weight quantization \cite{Bai2018,Hubara2016}, knowledge distillation \cite{Hinton2015,Tang2020} and network pruning \cite{Azarian2020,Han2015b}.
Network pruning has received by far the most attention and can be traced back to the 1990s when optimal brain damage \cite{Yann1990} and optimal brain surgeon \cite{Hassibi1993} were investigated. Both use second derivative information to assess the increase in the loss function of the network when a single weight is set to zero. In this way, the Hessian of the loss function is used to guide weight pruning, see also \cite{Liu2018} for a discussion. 

Further published methods are often based on complex \emph{pruning criteria} which decide which weight, neuron or filter is set to zero. Examples are the Hessian of the loss function used in \citet{Lecun1998}, the first and second-order Taylor expansions needed in \citet{Molchanov2019} or the entropy calculated in \citet{GEOKSEE2003}. Their evaluation requires a significant part of the additional calculation costs, most of them at least in the order of a gradient computation as analyzed by \citet{Yeom2019}. Consequently, the effort to find the most relevant connections in the network before or after the actual training is enormous, not to mention the often time-consuming search for suitable hyperparameter settings. This is in contrast to the paradigm of automatic network architecture design, which is denoted as automated machine learning (\emph{AutoML}). Simpler pruning criteria for individual weights have been established which include approaches based on the $l_1$-norm of weights, see, e.g., \cite{Han2015b,Yang2019}. 

%2. neuste forschung über die verschiedenen pruning strategies 
Pruning methods can be categorized w.r.t.\ two different pruning strategies. In \emph{structured pruning} neurons or entire filters (i.e., complete substructures of the network) are removed, see, e.g., \cite{He2018,Kuzmin2019,Mummadi2019}.
In contrast to this, \emph{unstructured pruning} aims at removing individual weights that are labeled as less relevant, see, e.g., \cite{Azarian2020,Yang2019,Zhang2020,Zhang2018}. The two approaches usually lead to quite different types of weights: While in the first case, pruning neurons or feature maps implies that complete tensors are dropped, resulting in dense weight vectors of reduced dimension, pruning single edges in the latter case leads to sparse weight vectors that retain the original dimension.

Most of the approaches use pruning in a-priori learning strategies that aim at finding a best possible starting architecture. This is called the \emph{winning ticket} in \cite{Frankle2018}. It can also be integrated during or after the training, some approaches even prune and retrain the weights many times \cite{Han2015}. In most cases, weights or neurons which have been set to zero once during the training will be excluded from further consideration, which is referred to as \emph{hard pruning} \cite{Han2015}.
A \emph{soft pruning} approach is suggested, e.g., in \cite{Guo2016} where pruned weights might exceed a given threshold value later during the training and are hence reconsidered during the remaining training steps.
It can be observed that many pruning approaches increase the overall training time as a result of costly a-priori training and/or a-posteriori fine tuning, see, e.g., \cite{Frankle2018,Han2015}. Moreover, recent papers question the efficiency of pruning methods and disprove the efficiency of the winning ticket architecture by varying the learning rate in structured pruning methods, see, e.g., \cite{Liu2018}. 
Other authors interrelate network pruning with adversarial attacks \cite{Gui19,Ye2019}. It was empirically shown in \cite{Guo2018} that sparse models resulting from unstructured pruning lead to a higher robustness against adversarial attacks. 
Common to all approaches is that the relative importance of the pruning criterion has to be set manually depending on the given training data, the network architecture and the remaining hyperparameters. 

%MOP
Considering the network complexity as an equitable second optimization goal in a biobjective optimization model (rather than as a regularization term) 
is not new, see, e.g., \cite{yin06multi,Jin2008,Braga2006,Furukawa2006} and \cite{Teixeira2000}, in which evolutionary algorithms are used.
However, these methods usually use time-consuming a priori procedures and are not yet well established in the machine learning community. 
%4. sehr ähnliche 
In order to improve the performance of multiobjective training approaches we suggest to use a simple and efficient $l_1$-objective as a second optimization criterion that guides the pruning process. 
A similar approach was suggested in \cite{Lym2019}, where pruning is an integral part of the network training, however, without considering regularization as a method to guide the pruning process. While we aim at reducing both the cost of training and of inference, \cite{Lym2019} focuses primarily on reducing the cost of the training. 
Integrating pruning in the training process is also suggested for structured pruning by \cite{Wang2019}. To the best of our knowledge, \cite{Liu2019} is the only reference that extends SGD to multiobjective optimization. However, their SMGD algorithm has not yet been used for a biobjective neural network training with conceptually different objective functions, nor for training in a Keras and Tensorflow environment.

%5. Remainder of the paper 
\subsection{Contribution and Structure of the Paper}
The main contributions of this paper are (i) a consistent multiobjective perspective on two conflicting training goals in machine learning, (ii) a thorough review, implementation and testing of stochastic gradient descent algorithms in the context of multiobjective optimization, (iii) an intra-training pruning strategy to reinforce the reduction of the complexity of neural networks, and (iv) a novel stochastic dichotomic search approach to approximate favorable knee solutions on the Pareto front that is tailored for problems with fundamentally different objectives and when scalarized subproblems can not be guaranteed to yield globally optimal solutions.

The remainder of this paper is organized as follows: In Sections~\ref{sec:mult} and~\ref{sec:optim} we present the general modelling assumptions and the theoretical background. In Section~\ref{sec:mult} we take a multiobjective view on the trade-off between prediction accuracy and network complexity, and in Section~\ref{sec:optim} the underlying methods and optimization techniques from machine learning are reviewed.
Novel intra-training pruning strategies that are based on the biobjective re-interpretation of the regularized loss function are presented in Section~\ref{sec:pruning}. Moreover, approaches for finding knee solutions of the Pareto front of the biobjective model are discussed in Section~\ref{sec:knee}.
In Section~\ref{sec:exp} our numerical experiments are presented. All experiments are based on two widely used network architectures and datasets from image classification, namely MNIST \cite{lecun2010mnist} and CIFAR10 \cite{Krizhecsky2009}. We document the experiments for three different pruning methods and compare the success of pruning in dependence of different optimizers. The advantages and disadvantages of the stochastic multi-gradient optimization methods are illustrated and different approaches for approximating the Pareto front are compared. 
Finally, we summarize our results in Section~\ref{sec:conc} and mention future research directions. 
\section{A Multiobjective View on Training Neural Networks}\label{sec:mult}
%1. warum klassische loss funktionen alleine blöd
Classical objective (loss) functions in machine learning, for example, the cross entropy loss, the mean squared error or the log likelihood function, solely measure the performance of the neural network on the training data and are used to represent the prediction accuracy of the network. Prediction accuracy is defined as the ratio of the number of correct predictions to the total number of predictions.  We exemplarily consider  the cross entropy loss function for classification problems, which is to be minimized. For an arbitrary but fixed class (i.e., a fixed output category) the cross entropy is given by
 \begin{align}
E_{\text{CE}}(w,y^d)&= - \frac{1}{M}\sum_{i\in S^d} y^d(i) \cdot \log(y(i)) \label{eq:CE}.%\\
\end{align} 
Here, $S^d$ denotes the index set of \(M:=|S^d|\) training samples $(x^d(i),y^d(i))$, $i\in S^d$. Moreover, $y^d(i)$ denotes the given correct binary classification of the $i$-th sample $x^d(i)$, i.e., $y^d(i)=1$ if the $i$-th sample is in the considered class and $y^d(i)=0$ otherwise. Corresponding to this, $y(i)\in(0,1)$ is the probability determined by the network for the $i$-th sample being in the considered class. Assuming a predetermined network structure, these probabilities depend on the decision variables $w\in\R^N$ (weights of the neural network) and that $y^d$ is defined on a probability space (with probability measure independent of $w$) for which we assume that i.i.d.\ samples can be observed and generated. We refer to \cite{goodfellow16deep} for a more detailed and comprehensive introduction to neural networks.

Typically, DNNs with large degrees of freedom achieve excellent prediction accuracies w.r.t.\ the training data (denoted by $(x^d,y^d)$). However, the generalization, i.e., the performance on \emph{test data} (denoted by $(x^t,y^t)$), is often unsatisfactory due to overfitting. Overfitted models are also prone to adversarial attacks and thus their applicability is limited, see, e.g., \cite{Guo2018}. Indeed, we usually do not only aim at an excellent performance on the training data, but rather envisage a versatile network that generalizes well to various datasets whose characteristics are not necessarily known beforehand. 

%3. die zweite Zielfunktion: der reularisierungsterm 
An effective approach to avoid overfitting is to extend the loss function by a regularization term that, for example, penalizes large weights and/or minimizes the number of nonzero weights, thus trying to reduce the network complexity. The regularized  objective function can then be written as 
 \begin{equation}
  J^{\text{R}_{\lambda}}(w,y^d) = E(w,y^d) + \lambda\cdot\Omega(w) \label{eq:regularization}
 \end{equation} 
with a loss function $E$ %(for example, the cross entropy \eqref{eq:CE}) 
and a regularization term $\Omega$ that measures  the complexity of the network. The regularization hyperparameter $\lambda>0$ in \eqref{eq:regularization} reflects the relative importance of regularization and is of central importance for the quality of the trained network. Examples for regularization terms are \\ the weight decay ($l_2$-regularization)
\begin{equation}
\Omega_{l_2}(w)=\frac{1}{2}\sum_{n=1}^N w_n^2, \label{eq:l2}
\end{equation} 
the sum of absolute values of weights ($l_1$-regularization) 
%\vspace{-1ex}
 \begin{equation} \Omega_{l_1}(w)=\sum_{n=1}^N |w_n|,\label{eq:l1}
 \end{equation} 
%\vspace{-2.5ex}
and the number of nonzero weights ($l_0$-regularization)
%\vspace{-1ex}
 \begin{equation} \Omega_{l_0}(w)=\sum_{n=1}^N \delta(w_n), \label{eq:l0}
 \end{equation} 
with $n$ being an index summing over all weights of the neural network, and $\delta(w_n)=1$ if and only if $w_n\neq0$.  Observe that the $l_0$-regularization \(\Omega_{l_0}\) is only piece-wise differentiable on intervals on which it has a constant value, i.e., its gradient is zero. Consequently, \(\Omega_{l_0}\) can not directly be included as a regularization term in a gradient-based training approach. 

%4. neu interpretiert
In the following, we re-interpret the loss function and the regularization function as two \emph{independent} training goals. The regularization hyperparameter $\lambda$ then reflects the \emph{trade-off} between prediction accuracy and network complexity.
In other words, we want to \emph{simultaneously} optimize the prediction accuracy \emph{and} the network complexity in a biobjective model
 \begin{equation}
\min_{w\in \R^N} J(w,y^d)=(J_1(w,y^d), J_2(w))= \bigl(E(w,y^d),\Omega(w) \bigr).\label{eq:biobjective}
 \end{equation} 
Using mathematical methods of multiobjective optimization now offers a new perspective on neural network training and thus paves the way for an adaptive decision support tool for preferable network architectures.

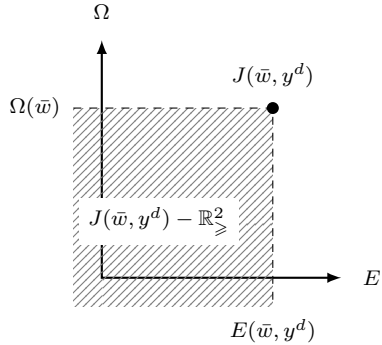
\begin{figure}[htbp]
  \centering
  \begin{tikzpicture}[scale=1.5]
   \usetikzlibrary{patterns}
      % Draw axes
      \draw [latex-latex,thick] (0,2.1) node (yaxis) [above=1ex] {\footnotesize$\Omega$}
	  |- (2.1,0) node (xaxis) [right=1ex] {\footnotesize $E$};
	  % Draw two intersecting lines
  %    \draw (0,0) coordinate (a_1) -- (2,1.8) coordinate (a_2);
  %    \draw (0,1.5) coordinate (b_1) -- (2.5,0) coordinate (b_2);
      % Calculate the intersection of the lines a_1 -- a_2 and b_1 -- b_2
      % and store the coordinate in c.
      \coordinate (c) at (1.5,1.5);
      % Draw lines indicating intersection with y and x axis. Here we use
      % the perpendicular coordinate system
      \pattern[gray, pattern=north east lines, pattern color= gray] (-.25,-0.25) rectangle (1.5,1.5);
      \node[fill=white] at (0.5,0.5) {\footnotesize\(J(\bar{w},y^d)-\mathbb{R}^2_{\geqslant}\)};
      \draw[dashed] (-0.25,1.5) node[left=0.3ex] {\footnotesize $\Omega(\bar{w})$}
	  -| (1.5,-0.25) node[below=0.3ex] {\footnotesize $E(\bar{w},y^d)$};
      % Draw a dot to indicate intersection point
      \fill[black] (c) circle (1.5pt) node[right=1ex,above=1ex] {\footnotesize $J(\bar{w},y^d)$};
  \end{tikzpicture}
  \caption{Illustration of the outcome $J(\bar{w},y^d)\in\mathbb{R}^2$ of a particular weight $\bar{w}$ in the objective space of problem \eqref{eq:biobjective}. \label{fig:objectivespace}}
\end{figure}

Fig.~\ref{fig:objectivespace} illustrates one possible outcome vector in the objective space of problem \eqref{eq:biobjective}. 
%6. Pareto optimal and nondominted
Assuming that the training data $(x^d,y^d)$ is given and fixed, a weight vector
$\bar{w}\in \R^N$ is called \emph{Pareto optimal} if and only if 
\[\nexists \; w\in \R^N:\; J(w,y^d)\leqslant J(\bar{w},y^d),\]
i.e., there is no $w\in \R^N$ such that $E(w,y^d)\leq E(\bar{w},y^d)$ and $\Omega(w)\leq\Omega(\bar{w})$, where at least one of these two inequalities is strict. The corresponding image $J(\bar{w},y^d)$  is then called \emph{nondominated}. Thus, Pareto optimal weight vectors are such weight vectors that can not be improved w.r.t.\ both objective functions, $E$ and $\Omega$, simultaneously.
Note that the outcome vector  $J(\bar{w},y^d)$ illustrated in Fig.~\ref{fig:objectivespace} is nondominated if and only if the cone $J(\bar{w},y^d)-\mathbb{R}^2_{\geqslant}$ does not contain any other feasible outcome vector. We refer to \cite{ehrgott05multicriteria} for a comprehensive introduction to multiobjective optimization. 

%7. Pareto front and weighted sum
An approximation of the \emph{Pareto front}, i.e., the set of \emph{all} nondominated outcome vectors in the objective space, provides trade-off information and supports the network developer in selecting a most preferred network architecture.
Algorithmically this can be realized, for example, by using evolutionary multiobjective optimization (EMO) algorithms, see e.g., \cite{yin06multi}, with the disadvantage of possibly very high training costs depending on the size of the network and the training data. A simple and efficient alternative is the successive solution of parametric scalarizations, e.g., the weighted sum scalarization, which is given by an unconstrained single-objective optimization problem
 \begin{equation}
\min_{w\in \R^N} J_{\lambda}(w,y^d)= (1-\lambda)\cdot E(w,y^d)+\lambda \cdot\Omega(w) \label{eq:weightedsum}
 \end{equation} 
with a weighting parameter $\lambda \in [0,1]$. It is well known that an optimal solution of a weighted sum scalarization \eqref{eq:weightedsum} with $\lambda\in(0,1)$ is always Pareto optimal. Under convexity assumptions the weighted sum scalarization also guarantees to generate all Pareto optimal solutions by variation of the weighting parameter \(\lambda\in[0,1]\).

Note that problems \eqref{eq:regularization} and \eqref{eq:weightedsum}  are closely related. Indeed, minimizing the regularized objective function \eqref{eq:regularization} with a fixed regularization hyperparameter $\bar{\lambda}>0$ is equivalent to minimizing the weighted sum scalarization \eqref{eq:weightedsum} with weighting parameter $\tilde{\lambda}=\frac{\bar{\lambda}}{1+\bar{\lambda}}$. Conversely, every weighted sum scalarization \eqref{eq:weightedsum} with a fixed weighting parameter $\tilde{\lambda}\in [0,1)$ can be equivalently formulated using a regularized objective function \eqref{eq:regularization} with regularization hyperparameter $\bar{\lambda}=\frac{\tilde{\lambda}}{1-\tilde{\lambda}}$, while this is not possible when  $\tilde{\lambda}=1$. However, this difference is not important in the context of neural network training since minimizing $\Omega(w)$ without considering a loss function $E(w,y^d)$ usually leads to collapsed networks (with all weights equal to zero) which is not meaningful.
As a consequence, regularization can be interpreted as a particular weighted sum scalarization of the biobjective problem \eqref{eq:biobjective}, and hence an optimal solution of problem \eqref{eq:regularization} is always Pareto optimal for the biobjective problem \eqref{eq:biobjective}. We argue that the biobjective perspective supports the analysis of the trade-off between the ``original'' and the ``regularizing'' objective and hence supports an informed or even automatic choice of the weighting parameter $\lambda$. Moreover, it opens the door for using other scalarizations that can potentially find solutions on non-convex parts of the Pareto front.

One may observe a Pareto front forming a sharp \emph{knee}, a point on the Pareto front, where the trade-off significantly changes, see, e.g., \cite{das99}.
Such knee points are interesting since they often provide good compromise solutions, i.e., in our case they achieve a relatively high prediction accuracy with a relatively small number of nonzero weights.

In the following, we focus on a particular choice for $E$ and for $\Omega$, namely the cross entropy $E=E_{\text{CE}}$, see \eqref{eq:CE}, and the $l_1$-regularization $\Omega=\Omega_{l_1}$, see \eqref{eq:l1}. To reduce the number of nonzero weights we combine \(l_1\)-regularization with an intra-training pruning strategy which is described in detail in Section~\ref{sec:pruning}.

However, the concept is generally applicable and can also be implemented with other choices for the loss function $E$ and the regularization function $\Omega$.

\section{Stochastic Gradient Descent Algorithms used in ML}
\label{sec:optim}
\subsection{Stochastic Gradient Descent and Extensions} \label{subsec:singlegrad}
%1. was bisland an optimierern in ML vorrangig genutzt wird 
We first consider the single-objective optimization problem \eqref{eq:weightedsum} with loss function $J_1(w,y^d)=E(w,y^d)$ and regularization $J_2(w)=\Omega(w)$. %\eqref{eq:CEandl1}. 
The \emph{stochastic gradient descent (SGD)} algorithm, suggested already in 1951 by \citet{Robbins1951}, is probably the most popular and widely used optimization algorithm in machine learning. It avoids the high computational cost of computing the true gradient $\nabla_{\! w} J_1(w,y^d)$ of the loss function $J_1(w,y^d)$ (which grows with the size of the sample vector $y^d$) by considering only one -- or a \emph{mini-batch} of -- randomly chosen sample(s) from $y^d$ to approximate $\nabla_{\! w} J_1(w,y^d)$ in each iteration. 
When only one sample $i\in S^d$ is randomly chosen, then %For a random choice of one sample, $(x^d(i),y^d(i)) \in S $, 
the stochastic gradient of a loss function of the form $J_1(w,y^d)= \frac{1}{M}\sum_{i\in S^d}  J_1\bigl(w,y^d(i)\bigr)$ %in $w$ 
is defined as
%  g(w,y^d) %=\frac{\partial}{\partial w} J_1\bigl(w,y^d(\kk{j})\bigr)
\begin{equation*} 
  g(w,y^d)=  \nabla_{\! w} J_1\bigl(w,y^d(i)\bigr).
\end{equation*}   
Similarly, for a mini-batch, i.e., for a random subset $P \subset S^d $ %, $(x^d(i),y^d(i)) \in P \; \forall \; i \in \{1,\ldots,p\}$ (on a mini-batch) 
with batch size $p= \vert P \vert \leq M$, it is given by
%= \frac{\partial}{\partial w} \, \frac{1}{p} \sum_{\ms{j\in P}} J_1\bigl(w,y^d(j)\bigr) 
\begin{equation} \label{eq:stochgrad}
  g(w,y^d)=\frac{1}{p}\sum_{i\in P}   \nabla_{\! w} J_1\bigl(w,y^d(i)\bigr).
\end{equation}   

In combination with stochastic gradients the gradient descent algorithm becomes very efficient and useful for large-scale optimization. Note that the regularization objective $J_2(w)=\Omega(w)$ does not depend on the training data, and hence its true gradient $\nabla_{\! w} J_2(w)$ can usually be computed efficiently. 
The use of mini-batches is motivated by the fact that only the expectation of the stochastic gradient is a descent direction for a loss function at a given solution $w$ and the performance of the algorithm is sensitive to the variance of the stochastic gradient. Other variance reduction techniques are discussed, e.g., in \cite{bottou2018}.

To simplify the notation, we denote the weight vector of the $k$-th iteration by $w^k$ and the stochastic gradient \eqref{eq:stochgrad} at $w^k$ by $g^k:=g\bigl(w^k,y^d\bigr)$. Furthermore, let $w^0$ denote the (randomly chosen) initial weight vector and let the sequence of learning rates (i.e., step sizes) be given by $(t_k)_{k\in\mathbb{N}}$. A typical stopping criterion in machine learning is the number of \emph{epochs} $\kappa$ that counts how often each training sample is taken into account. Note that the batch size and the size of the training data hence defines the number of iterations in each epoch as the random choice in the stochastic gradient computation is implemented without replacement. The complete SGD algorithm  is summarized in Algorithm~\ref{alg:sgd} for the minimization of a loss function $J_1$. An extension to handle weighted sum objectives \eqref{eq:weightedsum} is immediate. We refer to \cite{sacks1958} for a convergence analysis.  

%%%%%%%%%%%%%%%%%%%-----SGD-----%%%%%%%%%%%%%%%%%%%%%%%%%%%
% k läuft über iterationen (für jede theoretisch eine neue lernrate möglich) 
% n läuft über die einzelnen Gewichte bis N(anzahl gewichte)
% j läuft über die Anzahl der Objectives bis m (anzahl der zielfunktionen)
% i läuft über die samples in S  bis batchsize p oder sample size M
\begin{algorithm}
 \caption{(Mini-batch) Stochastic Gradient Descent (SGD)}\label{alg:sgd}
  \KwIn{Training objetive $J_1$, training data $S^d$, learning rate sequence $(t_k)_{k\in\mathbb{N}}$, random initial weights $w^0$, stopping criterion, batch size $p$, $k=0$}
  \KwOut{Trained model parameters $w^*$}
  %$w^0 \leftarrow w^\sigma $ \;
  %$ k=0$\;  
  \Repeat{stopping criterion}{$S \leftarrow S^d$\;
    \While{$|S|\geq p$}{
    randomly choose $P \subseteq S$ with $|P|=p$\;
    $g^k\leftarrow \frac{1}{p} \sum_{i\in P} \nabla_{\! w} J_1 (w^k,y^d(i))$\;
    \For{$n \in \{1,\ldots,N\}$}
    {
      $w^{k+1}_n \leftarrow %w^{k}_n - t_k \cdot (\frac{1}{p} \sum_{\ms{i\in P}} \frac{\partial}{\partial w_n}\, J_1\bigl(w^k,y^d(i)\bigr)= 
      w^{k}_n - t_k \cdot g^k_n$\;
   %\hl $= w^k_n - t_k (\frac{\partial}{\partial w^k_n} \frac{1}{p}\sum_{i=1}^p (E_{\text{CE}}(w^k,y^d(i))+\lambda \,\Omega_{l_1}(w^k) ) $\;
    } 
    $k\leftarrow k+1$\;
    $S \leftarrow S \setminus P$\;
    }
  }
\end{algorithm}
%2. was so für standard hyperparmeter gesetzt werden können
Most frequently, a constant learning rate $t_k=c \in (0,1]$ (for all $k$) is used in neural network training. However, it is often advisable to integrate a decreasing (or oscillating) sequence of adjustable learning rates. We refer to a predefined sequence of  decreasing learning rates as a \emph{learning rate schedule (LRS)}. The Keras library \cite{chollet2015keras} comes with a time-based LRS which is controlled via a $\decay$ hyperparameter. Then the learning rate in iteration $k$ is computed using the formula
  \begin{equation} 
 t_k= t_0 \cdot \frac{1}{1+\decay \cdot k} . \label{eq:LRS}
\end{equation}
In addition, the SGD implementation of Keras offers the possibility to use a \emph{momentum} term. In this case it requires an additional momentum hyperparameter $M_O$. Then the momentum  $v^k$ in iteration $k$ is recursively calculated, starting from the initial value $v^0= 0$, as
\begin{equation*}
  v^k = v^{k-1}\cdot M_O - t_k \cdot g^k,  
\end{equation*}
and the weight update in line~$7$ of Algorithm~\ref{alg:sgd} changes to
\begin{equation}
  w^{k+1} \leftarrow w^{k} + v^k. \label{eq:mom}
\end{equation}
SGD with momentum usually converges faster than the standard algorithm. Indeed, the momentum method helps to follow prevalent descent directions and reduces oscillation caused by the variance in the stochastic gradients. Therefore, we can use a higher learning rate when using momentum. A typical choice for the $\decay$ is $0.01$. For the momentum hyperparameter it is %empirically 
recommended to choose $M_O \in [0.5,0.9]$.
 
%3. extensions of SGD
Several variants and extensions of vanilla SGD implementations are available. Popular examples are the \emph{RMSProp} \cite{GHinton} and the \emph{Adam} \cite{kingma14adam} optimizer. The differences between these variants relate to the calculation of the weight updates which also influence the effect of the learning rates during the training. To better understand their generalization towards solving multiobjective problems, i.e., SMGD algorithms, some more details are described below. 

The central idea of \emph{RMSProp} is to use moving averages of the squared components of the stochastic gradients to scale the step lengths, which helps to smoothen the search and again avoids overly large oscillations of the lengths of the actual steps.
In this context, the (overall) step size can be interpreted as an adaptive learning rate that changes over time.
%%%%%% RMSProp
To implement the RMSProp algorithm in the context of the vanilla SGD algorithm (Algorithm~\ref{alg:sgd}), the following line has to be added directly after line $6$ of Algorithm~\ref{alg:sgd}:
 \begin{equation}
\text{mov}^k\bigl((g^k_n)^2\bigr)= \beta \cdot \text{mov}^{k-1}\bigl((g^{k-1}_n)^2 \bigl) + (1-\beta) \cdot (g^k_n)^2,
 \end{equation}\label{eq:movk}
where $\beta\in(0,1)$ is a hyperparameter that balances between the current and the previous iteration.
Then line $7$ of Algorithm~\ref{alg:sgd} changes to
 \begin{equation}\label{eq:rmsprop}
w^{k+1}_n \leftarrow w^{k}_n - \frac{t_k\, g^k_n}{\sqrt{\text{mov}^k((g_n^k)^2)}} % \Bigl(\frac{1}{p} \sum_{\ms{j\in P}} \frac{\partial}{\partial w_n} \, J_1(w^k,y^d(j)\Bigr).
 \end{equation}

Note that using a momentum hyperparameter of $M_O$ in the vanilla SGD actually has a similar effect. 

The optimization algorithm \emph{Adam} %\cite{kingma14adam} 
is based on adaptive estimates of first and second order moments of the stochastic gradients. In this way, individual learning rates are calculated for \emph{each} component $w^k_n$ of the weight vector $w^k$ (in iteration $k$). Adam uses squared components of the stochastic gradients like RMSProp to scale the learning rate and uses momentum with moving averages of the gradient. Consequently, it can be interpreted as a combination of RMSProp and SGD with momentum, and it is is well comparable with SGD when both momentum and learning rate schedules are applied. 
A specific advantage of Adam is that the magnitudes of the updates of the weights are invariant to a re-scaling of the stochastic gradient.  %Therefore, it balances differently weighted outputs of the neural networks with respect to each weight. 
Moreover, the learning rates are limited by the step size hyperparameter which implies an implicit annealing of the learning rate when updating the weights. % in line $6$ of Algorithm~\ref{alg:sgd}.
In the vanilla SGD algorithm (Algorithm~\ref{alg:sgd}) this requires the following extensions after line~5: 
\begin{equation}\label{eq:moment}
\begin{aligned}
m^k_n = \beta_1\cdot m^{k-1}_n + (1- \beta_1) \cdot g^k_n \\ 
v^k_n = \beta_2 \cdot v^{k-1}_n + (1- \beta_2) \cdot (g^k_n)^2 \\
\hat{m}^k_n =  \frac{m^k_n}{1-(\beta_1)^k} \\ 
\hat{v}^k_n = \frac{v^k_n}{1-(\beta_2)^k},
\end{aligned}\end{equation}
where $\beta_1,\beta_2\in (0,1)$ and $\epsilon=10^{-7}$ (to avoid dividing by zero). 
Line $7$ of Algorithm~\ref{alg:sgd} then changes to
\begin{equation}\label{eq:adam}
 w^{k+1}_n \leftarrow w^{k}_n - t_k \cdot \frac{\hat{m}_n^k}{\sqrt{\hat{v}_n^k } + \epsilon} .
\end{equation}
We refer to \cite{bottou2018} for a more detailed introduction to single objective descent algorithms used for neural network training. 

\subsection{Stochastic Multi-Gradient Descent and Extensions}\label{subsec:multigrad}
%1. multiobjective 
Instead of first converting the biobjective problem \eqref{eq:biobjective} into one (or a series of) single-objective scalarizations, e.g., using weighted sum scalarizations \eqref{eq:weightedsum}, we now consider a multiobjective gradient descent algorithm for neural network training. 

Deterministic multiobjective gradient descent algorithms were first introduced by \cite{fliege2000steepest}. Starting from an initial solution, the idea is to iteratively move into directions that are steepest \emph{common} descent directions, and hence descent directions for \emph{all} objective functions. For an unconstrained and continuous problem, the algorithm terminates with a stationary point (a local Pareto optimal solution) when no such direction exists. Multiobjective gradient descent algorithms do not require any preference information and generally converge quickly (under appropriate assumptions)  to \emph{one} local Pareto optimal solution. However, guiding the search towards specific regions of the Pareto front without prior knowledge (e.g., towards a knee solution) is generally difficult.

An extension of this method to problems where, rather than true gradients, only stochastic gradients can be efficiently computed was suggested in \cite{Liu2019}. 
The \emph{stochastic multi-gradient descent} (SMGD) algorithm is based on earlier works by \cite{Caballero2004} and \cite{Desideri2018}. \citet{Liu2019} provide a convergence analysis that is similar to that for the SGD algorithm.
They specifically apply the SMGD algorithm for the detection of dependencies in training data sets, but actually not for neural network training itself.

The SMGD algorithm is defined for general multiobjective optimization problems of the form $\min_{w\in\R^N} J(w,y^d)=  \bigl(J_1(w,y^d),\ldots,  J_q(w,y^d) \bigr)$, where $q\geq 2$ denotes the number of objective functions. In our case, we have $q=2$ and the second objective function $J_2(w,y^d)=J_2(w)=\Omega(w)$ does not depend on the training data. As a consequence, the stochastic gradient $g_2(w,y^d)=\nabla_{\! w} J_2(w)$ of $J_2$ coincides with the true gradient and can be efficiently computed. To keep the notation simple when describing the SMGD algorithm, we will nonetheless refer to both gradients as ``stochastic gradients'' and adopt the notation $J_j(w,y^d)$ and $g_j(w,y^d)$, $j=1,2$. Similar to Section~\ref{subsec:singlegrad} we abbreviate $g^k=g(w^k,y^d)$ and $g^k_j=g_j(w^k,y^d)$. 
A common stochastic descent direction in iteration $k$ is then determined  as a weighted sum of the individual stochastic gradients
\begin{equation*}
g^k=\sum_{j=1}^q \lambda_j^k \cdot g_j^k ,
\end{equation*}
however, not with fixed weighting parameters as in a weighted sum approach but rather with variable weighting parameters $\lambda^k\in\R^q_{\geqq}$, $\sum_{j=1}^q \lambda_j^k=1$, that are selected such  as to obtain a \emph{steepest} common descent direction in each iteration.  %Note that only $q-1$ weighting paramters need to be specified since the last weighting parameter follows from the normalization constraint $\sum_{j=1}^q \lambda_j^k=1$. 
Note that when $q=2$ we equivalently set $\lambda_2=\lambda$ and  $\lambda_1=1-\lambda$ with only one weighting parameter $\lambda\in[0,1]$, see \eqref{eq:weightedsum}. 

The steepest common descent direction is determined by computing the weighting parameters $\lambda^k$ in iteration $k$ as an optimal solution of a convex quadratic optimization subproblem, i.e.,

\begin{align} \label{eq:subprob} 
\lambda^k = &\argmin  \; \biggl\Vert \sum_{j=1}^q \lambda_j\, g_j^k \biggr\Vert ^2 \\ 
&\text{ s.t. }   \;\lambda \in \Bigl\{ \lambda \in \R^q \colon \sum_{j=1}^{q} \lambda_j =1, \lambda_j \geq 0, \; \forall \; j \in \;\{ 1,\ldots,q\} \Bigr\} .\nonumber
\end{align}

For a more detailed derivation and theoretical foundation of stochastic multi-gradient descent we refer to \cite{Liu2019}.

%%%%%%%%%%%%%%%%%%%-----SMGD-----%%%%%%%%%%%%%%%%%%%%%%%%%%%
% k läuft über iterationen (für jede eine neue lernrate?!) 
% n läuft über die einzelnen Gewichte bis N (anzahl gewichte)
% j läuft über die Anzahl der Objectives bis q (anzahl der zielfunktionen)
% i läuft über die samples in S  bis p
\begin{algorithm}[htb]
 \caption{(Mini-batch) Stochastic Multi-Gradient Descent (SMGD)}\label{alg:smgd}
  \KwIn{Training data $S^d$, learning rate sequence $(t_k)_{k\in\mathbb{N}}$, random initial weights $w^0$, stopping criterion, batch size $p$, set $k=0$}
  \KwOut{Trained model parameters $w^*$}
  \Repeat{stopping criterion}{$S \leftarrow S^d$\;
     \While{$|S|\geq p$}{
     randomly choose $P \subset S$ with $|P|=p$\;
     $g_j^k\leftarrow \frac{1}{p} \sum_{i\in P} \nabla_{\! w} J_j (w^k,y^d(i))$ $\forall j \in \{1,\ldots,q\} $\;
    solve the subproblem \eqref{eq:subprob} to obtain $\lambda^k\in\R^q$\;
    $g^k \leftarrow \sum_{j=1}^q \lambda_j^k \cdot g_j^k$ \; 
    \For{$n \in \{1,\ldots,N\}$}
    {
    $w^{k+1}_n \leftarrow w^{k}_n - t_k \cdot g_n^k$\;
    }
$k \leftarrow k+1$\;
$S \leftarrow S \setminus P$\;
}
}
\end{algorithm}

The vanilla SMGD algorithm formulated in Algorithm~\ref{alg:smgd} can now be extended similar to the vanilla SGD algorithm given in Algorithm~\ref{alg:sgd} in order to improve the training results. We suggest two such variants of SMGD, namely a \emph{multiobjective RMSProp optimizer} (MRMSProp) and a \emph{multiobjective Adam optimizer} (MAdam).
Note that in our case the two objective functions $J_1=E_{\text{CE}}$ and $J_2=\Omega_{l_1}$ are fundamentally different, which will become apparent in the numerical tests presented in Section~\ref{sec:exp}. To avoid the risk that one of the objectives consistently overrides the other,  we hence suggest to compute the moving averages $\text{mov}^k_j$, $j=1,\dots,q$ for the MRMSProp algorithm according to \eqref{eq:movk} \emph{independently} for each of the $q$ objective functions and combine them using the weights $\lambda^k$ only when updating the weight vector according to \eqref{eq:rmsprop}. Similarly, we suggest to compute the estimates of the first and second moment for the MAdam optimizer according to \eqref{eq:moment} \emph{independently} for each objective function and combine them using the weights $\lambda^k$ only for the weight update according to \eqref{eq:adam}.   
As a consequence, the two goals of minimizing the prediction error and minimizing the network complexity are treated equivalently in our setting.

\section{Unstructured and Soft Intra-Training Pruning}\label{sec:pruning}
% 1. pruning strategy in general
To reinforce the effect of the second objective function $J_2(w)=\Omega_{l_1}(w)$ 
descent steps are complemented by an associated \emph{intra-training pruning} (ITP) strategy. 
Thereby, we focus on the weights of the dense layers of the neural network. Such pruning approaches are referred to as \emph{unstructured} pruning. Weight pruning is motivated by the observation that small weights have only a limited impact on the prediction accuracy of the neural network. By setting all weights to zero as soon as their absolute value falls below a predetermined \emph{threshold value} $\tau>0$, the corresponding edges are removed from the neural network which helps to reduce the network complexity. In our approach pruning is directly incorporated in the training and is guided by the second objective function $\Omega_{l_1}(w)$. 

We compare two different variants of the ITP strategy, namely batchwise and epochwise pruning, with after training pruning where only one single pruning step is applied after the neural network is fully trained. The most promising pruning strategy is then selected to be used as a baseline when comparing other algorithmic components. 
Note that we already start with a sparse initialized weight matrix in all pruning approaches, i.e., there is already one pruning step before the training starts.

% 2. two different strengtths/variants in more detail
In \emph{batchwise} pruning, pruning is applied after each batch, while in \emph{epochwise pruning}, a pruning step is performed after each epoch. In our implementation we use Keras callback functions to update the weight matrices of all dense layers. As ITP always ends with a pruning step after the network is totally trained, these methods can be seen as extensions of after-training pruning. It is important to note that in our approach, in contrast to, e.g., \cite{Han2015}, pruned weights are set to zero but dot not necessarily stay fixed at zero. Depending on the subsequent iterations weights may exceed the threshold and become relevant again. Hence, our strategy belongs to the class of \emph{soft} pruning methods. 
Thereby pruning is an integral part of the training and thus there is no need for a recovery phase as suggested by \cite{Zhu2017}, or other fine tuning methods.

% 3. erklärung
We argue that when using ITP then the network \emph{learns} that weights will be pruned if they become too small. The iterative pruning of small weights leads to an increasing number of zero weights. In general, not all zero weights are restored to values above the threshold level in subsequent iterations since this would significantly deteriorate the \(l_1\)-regularization term as stated in \cite{Lym2019}. Consequently, a smaller number of non-zero weights has to retain the information. It turns out that to a certain degree this is indeed possible without significantly deteriorating the prediction accuracy. This is confirmed by the numerical experiments presented in Section~\ref{sec:exp}.

A careful choice of the pruning threshold is important when fine-tuning the pruning strategy. Indeed, this hyperparameter also trades-off between prediction accuracy and network complexity. 
Our tests, however, show that fixed moderate threshold values in combination with varying weighting parameters for the $\Omega_{l_1}$ objective function effectively control the number of nonzero weights. While previous approaches search for an appropriate penalty value, making it expensive to include pruning from the beginning of the training, we fix this threshold beforehand. In our tests the threshold value varies between $\tau=0.001$ and $\tau=0.002$. Moreover, we observed that the learning rate has a strong impact on the pruning ratio and on the possibility that an individual weight exceeds the given threshold after being once set to zero.

\section{Finding Knee Solutions} \label{sec:knee} 

Knee solutions on the Pareto front are particularly interesting since they often realize a favorable trade-off between the considered objective functions. When training neural networks w.r.t.\ the two conflicting goals $J_1(w,y^d)=E_{\text{CE}}(w,y^d)$ (measuring prediction accuracy) and $J_2(w)=\Omega_{l_1}(w)$ (measuring network complexity) we observed pronounced knee solutions. Indeed, it turns out that a large number of weights can be set to zero without loosing much  prediction accuracy. An automatic detection of such knee solutions is highly desirable, particularly when aiming at AutoML approaches. 

However, SMGD methods like MAdam and MRMSProp as introduced in Section~\ref{sec:optim} are designed for a fast convergence towards the Pareto front, no matter where. Preference information can hardly be incorporated so that it is extremely unlikely that the final outcome is close to a knee solution (unless the methods are initialized with weights $w^0$ that are already close to a knee). This can also be observed in our numerical experiments, see Section~\ref{sec:exp}.

In order to nonetheless approximate knee solutions we develop scalarization-based approximation algorithms, namely a stochastic dichotomic search and a bisection search algorithm. Both methods aim at an efficient and automatic identification of weighting parameters (i.e., trade-offs) of the weighted sum scalarization \eqref{eq:weightedsum} that lead to knee solutions. 

The methods are developed to cope with the following challenges: (i) the (scalarized) optimization problems are truly large-scale, (ii) the two objective functions have fundamentally different characteristics (e.g., w.r.t.\ complexity, scaling, slope, curvature, and number of local minima), and (iii) the scalarized subproblems can generally not be solved to optimality. Indeed, when repeatedly training the same weighted sum scalarization with a single-objective variant of an SGD algorithm, we usually obtain different solutions that may be of largely differing quality. This motivates the introduction of a stochastic component in the usually deterministic selection of weighting parameters in a dichotomic search algorithm \cite{aneja1979}:  In Algorithm~\ref{alg:dicho} we suggest to make a random choice for the weighting parameter whenever the algorithm gets stuck which is recognized when weighting parameters are chosen too close to each other.

\begin{algorithm}[htb]
 \KwData{Training data $S^d$, hyperparameter settings for single-objective SGD solver, depth of the search (levels), initial weighting parameters $\lambda_1,\lambda_2\in[0,1)$, $\lambda_1<\lambda_2$, to approximate extremal solutions focusing on $J_1$ and $J_2$, respectively}
 \KwResult{Trained model parameters $w^*$ approximating a Pareto knee}
  $\mathtt{list} \leftarrow \{ \lambda_1, \lambda_2\}$\;
  $\mathtt{cand} \leftarrow \emptyset$\;
 \For{$l\in\mathtt{levels}$}{
    \For{$\lambda \in \mathtt{list}$}{
    train with weighted sum objective $J_{\lambda}$ (possibly with reduced epochs) and add objective vectors to $\mathtt{cand}$\;
    delete all dominated points in $\mathtt{cand}$\;
    }
 sort $\mathtt{cand}$ by second objective function (in increasing order)\;
 \If{$l<|\mathtt{levels}|$}{
 \For{$i\in\{2,\ldots,\vert \mathtt{cand}\vert\}$}{
 $\text{diff} \leftarrow \mathtt{cand}(i) - \mathtt{cand}(i-1)$\;
 $\lambda_{\text{new}} \leftarrow \frac{-\text{diff}_1}{\text{diff}_2 -\text{diff}_1}$\;%\bigl(\text{diff}_2, -\text{diff}_1\bigr)$\;
  \If{$\exists\; \hat{\lambda}\in\mathtt{list}\,:\,|\hat{\lambda}-\lambda_{{\rm new}}|\leq0.001$}{  
 randomly choose $\tilde{\lambda}\in \bigl[\max\{0.9\cdot \hat{\lambda},\lambda_1\}, \min\{1.1\cdot\hat{\lambda},\lambda_2\}\bigr]$\;
$\lambda_{\text{new}} \leftarrow \tilde{\lambda}$\;}
   add $\lambda_{\text{new}}$ to $\mathtt{list}$ \;}
}}
identify the Pareto knee in $\mathtt{cand}$ by comparing trade-offs\;
 \caption{Stochastic Biobjective Dichotomic Search \label{alg:dicho}}
\end{algorithm}

We compare the stochastic dichotomic search method to a simple yet effective bisection search method on the weighting parameters, see, e.g., \cite{bekasiewicz2017}:  
Starting from an initial interval of considered weighting parameters, this interval is successively decomposed into sub-intervals of equal size. Relevant intervals for further decomposition are identified by comparing trade-offs.

We remark that both algorithms suffer from the time intensive training, i.e., the calculation of optimized objective values for fixed weighting parameters, as in each case the neural network needs to be re-trained from scratch. To speed up this process only a reduced number of training epochs is used for the training.
When an approximation of the Pareto front has been computed with either of these methods, then a knee solution can be identified by comparing trade-offs. 

\section{Experiments}
\label{sec:exp}
%%%%%%%%% Architekturen etc aus Pruning Kapitel rüber

\subsection{Experimental Setup}
%%%%%%%% MNIST und LeNet
We validate our approach on the network architecture LeNet-$5$ \cite{Lecun1998}, a CNN consisting of two convolutional layers followed by three fully connected dense layers. After each convolutional layer the feature maps are subsampled by an average pooling layer. For details see Tab.~\ref{fig:sum:mnist}. In all layers, except the output \texttt{layer3}, ReLU activation functions are used. In the final dense output \texttt{layer3}, a softmax activation is used on the whole layer to ensure that the output represents a probability distribution with $y(i) \in (0,1)$ for each output category.

\begin{table}[htbp]\caption{LeNet-$5$ and VGG-like network architecture.}
\centering
  \scalebox{0.75}{
 % \footnotesize\ttfamily
 \centering
  \subfigure[LeNet-$5$ architecture for MNIST.\label{fig:sum:mnist}]{
  \begin{tabular}{l@{\extracolsep{1ex}}l@{\extracolsep{1ex}}l}
    \hline
    \textbf{Layer (type)}        & \textbf{Output Shape}       & \textbf{\# Param}   \\\hline
    conv2d\_29 (Conv2D)          & (None, 26, 26, 6)  & 60        \\\hline
    average\_pooling2d\_29       & (None, 13, 13, 6)  & 0         \\\hline
    conv2d\_30 (Conv2D)          & (None, 11, 11, 16) & 880       \\\hline
    average\_pooling2d\_30       & (None, 5, 5, 16)   & 0         \\\hline
    flatten\_15 (Flatten)        & (None, 400)        & 0         \\\hline
    layer1 (Dense)               & (None, 120)        & 48120     \\\hline
    layer2 (Dense)               & (None, 84)         & 10164     \\\hline
    layer3 (Dense)               & (None, 10)         & 850       \\\hline
    \multicolumn{2}{l}{Total params: 60,074} \\
    \multicolumn{2}{l}{Trainable params: 60,074} \\
    \multicolumn{2}{l}{Non-trainable params: 0} \\ \hline%\\
  \end{tabular} 
  }
  \hfill
  \subfigure[VGG-like architecture for CIFAR$10$.\label{fig:sum:cifar}]{
  \begin{tabular}{l@{\extracolsep{1ex}}l@{\extracolsep{1ex}}l}
    \hline
    \textbf{Layer (type)}    &\textbf{Output Shape} &\textbf{\# Param}    \\\hline   
    conv2d\_ 1 (Conv2D)      &(None, 32, 32, 32)    &896     \\\hline
    batch\_normalization\_1  &(None, 32, 32, 32)    &128     \\\hline
    conv2d\_ 2 (Conv2D)      &(None, 32, 32, 32)    &9248    \\\hline
    batch\_normalization\_2  &(None, 32, 32, 32)    &128     \\\hline
    conv2d\_ 3 (Conv2D)      &(None, 16, 16, 64)    &18496   \\\hline
    batch\_normalization\_3  &(None, 16, 16, 64)    &256     \\\hline
    conv2d\_ 4 (Conv2D)      &(None, 16, 16, 64)    &36928   \\\hline
    batch\_normalization\_4  &(None, 16, 16, 64)    &256     \\\hline
    conv2d\_ 5 (Conv2D)      &(None, 8, 8, 128)     &73856   \\\hline
    batch\_normalization\_5  &(None, 8, 8, 128)     &512     \\\hline
    conv2d\_ 6 (Conv2D)      &(None, 8, 8, 128)     &147584  \\\hline
    batch\_normalization\_6  &(None, 8, 8, 128)     &512     \\\hline
    denselayer (Dense)       &(None, 10)            &20490   \\\hline
    \multicolumn{2}{l}{Total params: 309,290} \\
    \multicolumn{2}{l}{Trainable params: 308,394}\\
    \multicolumn{2}{l}{Non-trainable params: 896}  \\\hline % \\
  \end{tabular}
  }}
  
\end{table}

In total, LeNet-$5$ consists of about $60,000$ trainable parameters (weights) of which approximately $59,000$ weights belong to the dense layers (\texttt{layer1}--\texttt{layer3}). 
We use \(l_2\)-regularization \eqref{eq:l2} in the convolutional layers and \(l_1\)-re\-gu\-lari\-zation \eqref{eq:l1} for the dense layers to reduce overfitting. While for the convolutional layer the \(l_2\)-regularization coefficient is fixed to $\tilde{\lambda}=0.0001$, we interpret the $l_1$-regularization for the dense layers as second objective function and vary the corresponding regularization hyperparameter $\lambda \geq 0$ %of the \(l_1\)-regularization for the dense layers 
to analyze its effect on pruning strategies. 
We train the network on the handwritten digits dataset MNIST \cite{lecun2010mnist} which does consist of $60,000$ labeled images defining the training data set $(x^d,y^d)$ and additionally $10,000$ labeled test images, stored as $(x^t,y^t)$.
No additional techniques (like dropout, data augmentation or batch normalization) are applied. In some of our tests we additionally apply momentum, learning rate schedules and/or weight decay as described in Section~\ref{subsec:singlegrad}. We train the model for $\kappa_{\text{max}}=30$ epochs and use, unless otherwise stated, the Adam optimizer with hyperparameters set to  the default values: 
$\beta_1=0.9$, $\beta_2=0.999$ and $\epsilon=10^{-8}$ and a constant learning rate of $t_k= 0.001$ for all $k$.

%%%%%% CIFAR und VGGG
We also examine the performance of ITP using a more complex network and the CIFAR$10$ dataset for image classification \cite{Krizhecsky2009}. The dataset consists of $60,000$ color images from $10$ classes, with $6,000$ images per class. The dataset is split into $\vert S^d \vert =50,000$ training data and $\vert S^t \vert =10,000$ test data.
The adapted VGG-like architecture \cite{simonyan14deep} is defined by six convolutional layers and one final \texttt{denselayer}, 
see Tab.~\ref{fig:sum:cifar}. Note that, for simplicity, only the layers including parameters are mentioned in the table. Additionally, there is a max pooling layer after each batch normalization layer. All convolutional layers use \emph{exponential linear units} (ELU) as activation functions, while the final \texttt{denselayer} uses softmax activation. In total, the VGG-like network consists of about $300,000$ trainable weights of which about $20,000$ belong to the final \texttt{denselayer}. We also use \(l_2\)-regularization with fixed coefficient $\tilde{\lambda}=10^{-6}$ on the convolutional layers and interpret the \(l_1\)-regularization on the \texttt{denselayer} as second objective function and vary the corresponding regularization hyperparameter $\lambda \geq 0$. 

\begin{figure}[tbh]
\centering
\scalebox{0.7}{
  \begin{tikzpicture}[>=stealth]
    \begin{axis}[
      width=9cm,
      height=5cm,
      scaled y ticks = false,
      y tick label style={/pgf/number format/fixed, /pgf/number format/fixed zerofill, /pgf/number format/precision=4},
      xmin=0,
      xmax=140,
      ymin=0,
      xlabel= epochs,
      ylabel= learning rate,
      label style={font=\footnotesize},
      tick label style={font=\footnotesize},
      ytick ={0,0.0001,0.0005,0.001},
      ]
      \addplot[thick,smooth] gnuplot [domain=0:160] {(-0.0009)*(exp((x-93.75)/(9.375))/(exp((x-93.75)/(9.375))+1))+0.001};
      \addplot[domain=0:139, gray, dashed] {0.0001};
      \addplot[domain=0:139, gray, dashed] {0.001};
    \end{axis}
  \end{tikzpicture} }
  \caption{Learning rate schedule reducing the learning rate over the $125$ training epochs.\label{fig:learningrate}}
\end{figure}
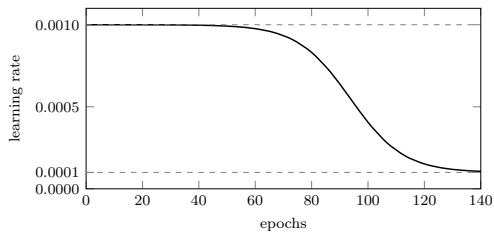

Since the CIFAR$10$ data set is considerably more complex than the MNIST data set we additionally apply classical techniques to get competitive results, including batch normalization after each convolutional layer and max pooling followed by dropout after the convolutional layers $2$,$4$ and $6$. 
Moreover, we use data augmentation in the training process and, unless stated otherwise, train with the RMSProp optimizer with hyperparameters set to the default values $\beta=0.9$ and $\epsilon=10^{-7}$ on $\kappa_{\text{max}}=125$ epochs. The learning rate is varied in different experiments. 
In particular, learning rate schedules (LRS) are used to reach a higher performance.
We develop an LRS that is specifically taylored for our methods and that smoothly reduces the learning rate over the epochs \(\kappa\in\mathbb{N}\), \(\kappa\in\{1,\ldots,\kappa_{\max}\}\), with a major drop after about \(75\%\) of the total number of epochs $\kappa_{\text{max}}$:
\begin{equation*}\label{eq:learningrate}
t(\kappa)= -( t_{\text{start}}-t_{\text{end}}) \,
\frac{\exp \bigl(\frac{\kappa - 0.75 \cdot \kappa_{\text{max}}}{0.075 \cdot \kappa_{\text{max}} } \bigr)}%
     {\exp \bigl(\frac{\kappa - 0.75 \cdot \kappa_{\text{max}}}{0.075 \cdot \kappa_{\text{max}} }\bigr) +1} +  t_{\text{start}}
\end{equation*} 
For CIFAR$10$ training with $\kappa_{\text{max}}=125$ epochs, and setting $t_{\text{start}}~=~0.001$ and $t_{\text{end}}~=~0.0001$,  the evolution of the learning rate is shown in Fig.~\ref{fig:learningrate}. Unless stated otherwise this schedule is used when talking about an LRS.

%%%%%%%%%%%%%%%% Ende Architekturen und LRS etc 
\subsection{Experimental Results and Trade-Off Analysis}
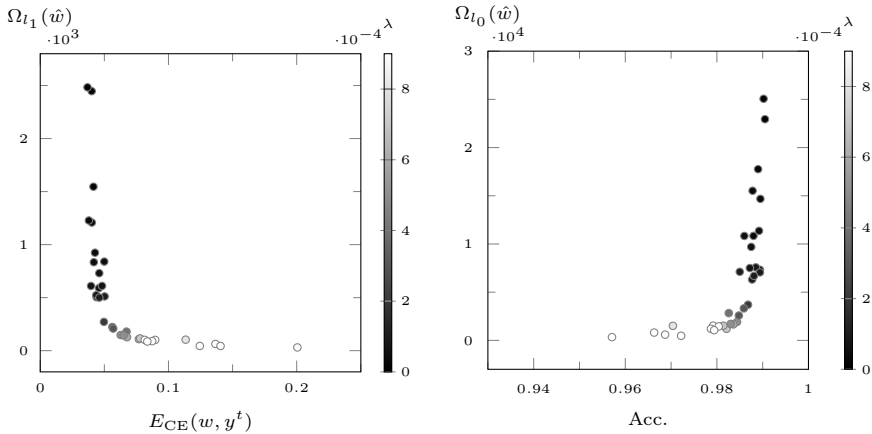
\begin{figure}[htb]
    \centering
    \subfigure[Cross entropy versus $l_1$-re\-gu\-la\-rization of the dense layers weights.\label{fig:CEversusL1(1.2)}]{
    \begin{tikzpicture}%[>=stealth]
      \begin{axis}[width=5.8cm, height=5.8cm, colorbar, 
        colormap={}{color(0cm)=(black);color(0.3cm)=(white)},
          colorbar style={ 
          font= \tiny,
          title= $\lambda$,
          width=0.1cm},
          point meta min=0,
          point meta max=0.0009,
	      ymin=-200,
	      ymax=2800,
	      xmin=0,
	      xmax=0.25,
	      minor tick num=1, 
	      ylabel={\(\Omega_{l_1}(\hat{w})\)},
	      xlabel={$E_{\text{CE}}(w,y^t)$},
	      x tick label style={yshift=-1pt},
	      y tick label style={xshift=-1pt},
	      scaled y ticks=base 10:-3,
	      every axis y label/.style={
		  at={(ticklabel* cs:1.05)},
		  anchor=south,
	      },
	      clip=true,
	      label style={font=\scriptsize}, 
	      tick label style={font=\tiny},
	      ]
 	      \addplot+[point meta=explicit, scatter src=explicit,scatter/use mapped color={
                 draw=gray,
                 fill=mapped color,
             }, scatter,mark = *, 
             only marks, 
             mark options={scale=0.7}]  table  [x index=5,y index=14,meta index=0, col sep=semicolon]{Test-final.txt};
	\end{axis}
    \end{tikzpicture}
    }
    \subfigure[Accuracy versus number of nonzero weights (\(l_0\)-regularization). \label{fig:L0versusAcc(1.2)}]{
    \begin{tikzpicture}
      \begin{axis}[width=5.8cm, height=5.8cm, 
      colorbar,  
      colormap={}{color(0cm)=(black);color(0.3cm)=(white)},
      colorbar style={
      font= \tiny,
       title= $\lambda$,
        width=0.1cm},
        point meta min=0,
    	point meta max=0.0009,
	      ymin=-3000,
	      ymax=30000,
	      xmin=0.93,
	      xmax=1,
	      minor tick num=1,
	      ylabel={\(\Omega_{l_0}(\hat{w})\)},
	      xlabel={Acc.\vphantom{\(\Omega_{l_1}(\hat{w})\)}},
	      x tick label style={yshift=-2pt},
	      y tick label style={xshift=-2pt},
	      every axis y label/.style={
		  at={(ticklabel* cs:1.05)},
		  anchor=south,
	      },
	      clip=true,
	      label style={font=\scriptsize}, 
	      tick label style={font=\tiny},
	      ]
      \addplot+[point meta=explicit, scatter src=explicit,scatter/use mapped color={
                draw=gray,
                fill=mapped color,
            },mark = *, 
            only marks, 
            scatter, 
            mark options={scale=0.7}]  table [x index=6,y index=13,meta index=0, col sep=semicolon]{Test-final.txt};
      \end{axis}
    \end{tikzpicture}
    }   
    \caption{LeNet-$5$ on MNIST dataset: Pareto fronts illustrating the trade-off between loss/accuracy and regularization for ITP.\label{fig:pareto(1.2)}}    
\end{figure}

We first report experimental results for our ITP strategy. To identify the most promising pruning approach, all training runs rely on the standard (single objective) SGD algorithms described in Section~\ref{subsec:singlegrad}. Subsequently, we present test results on the SMGD algorithms discussed in Section~\ref{subsec:multigrad}.
We observe that the effect of ITP varies epending on the value of the regularization hyperparameter $\lambda$ in \eqref{eq:regularization}. Indeed, the larger the impact of the \(l_1\)-regularization is, the more weights are pruned.
From a multiobjective perspective we evaluate different hyperparameter settings
including the threshold value $\tau$, the influence of the learning rate $t$ and of the regularization hyperparameter $\lambda$. Note that, although we use cross entropy \eqref{eq:CE} and \(l_1\)-regularization \eqref{eq:l1} as objectives (see Fig.~\ref{fig:CEversusL1(1.2)} for an example), we often use the \(l_0\)-regularization \eqref{eq:l0} and the prediction accuracy 
(see Fig.~\ref{fig:L0versusAcc(1.2)}) to indicate the success of ITP. 
However, both illustrations in Fig.~\ref{fig:pareto(1.2)} show the same behavior with a pronounced knee in the Pareto front.

Furthermore, we compare the SMGD algorithms that determine only one approximation of a Pareto optimal solution with scalarization based methods to approximate the Pareto front as introduced in Section~\ref{sec:knee}. Based on these approximations we identify knee solution. With $\lambda^*$ we refer to a marginal weighting parameter in \eqref{eq:weightedsum} corresponding to the knee solution. 
Note that, as the pruning strategy takes only dense layers into account, $\hat{w}$ refers to the weights in the dense layers of a network and, e.g., $\hat{w}^1$ denotes all weights in \texttt{denselayer1}. 

\subsubsection{Test Results for ITP on LeNet-$5$}\label{subsubsec:LeNet}
%1.was gemacht wurde
On the LeNet-$5$ network architecture %for the MNIST dataset s 
we compare \emph{batchwise pruning}, i.e., pruning weights below the threshold after each training batch, with \emph{epochwise pruning}, where pruning is only applied at the end of each epoch $\kappa$. The results are then compared with \emph{after training pruning}, where pruning is applied only at the end of the training (after all epochs). In all cases we initially prune the weight matrix before starting the training, and the pruning threshold is uniformly set to $\tau = 0.001$. 

\begin{figure*}[tbhp]
  \subfigure[$\Omega_{l_0}(\hat{w}^1)$ for the weights in \texttt{layer1} (\textit{batchwise ITP}).\label{fig:nonzeros:batchw}]{%
    \includegraphics[width=0.31\textwidth,trim=0 0 35 0,clip]{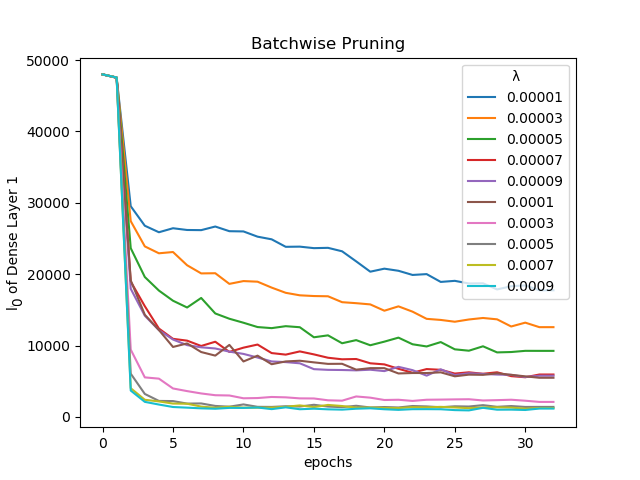}
  }
  \hfill
  \subfigure[$\Omega_{l_0}(\hat{w}^1)$ for the weights in \texttt{layer1} (\textit{epochwise ITP}).\label{fig:nonzeros:epochw}]{%
    \includegraphics[width=0.31\textwidth,trim=0 0 35 0,clip]{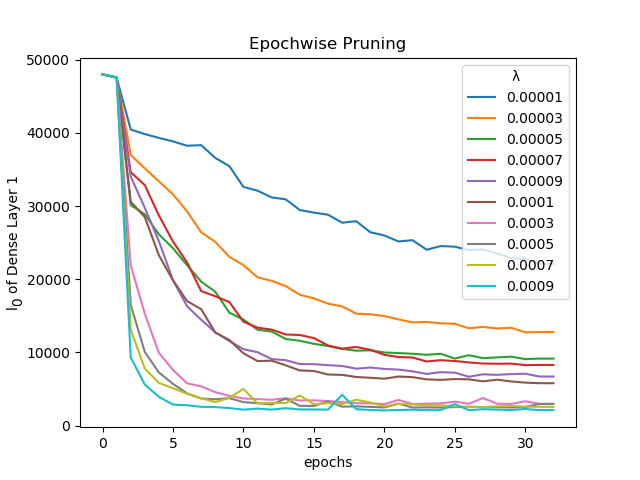}
  }
  \hfill
  \subfigure[Histogram of the weights in \texttt{layer3} (\textit{batchwise ITP}). \label{fig:violin}]{%
    \includegraphics[width=0.31\textwidth,trim=0 0 35 0,clip]{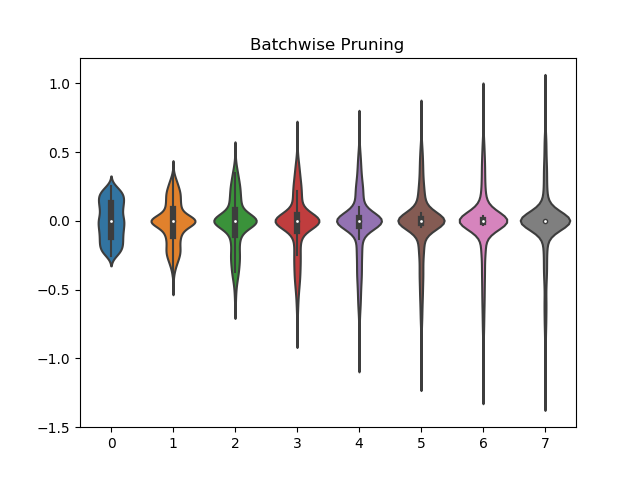}
  }
  \caption{LeNet-$5$ on MNIST dataset: Development of the regularization terms and the distribution of weights over the epochs of the training, in c) over every 5th epoch.}
\end{figure*}

%2.was die Ergebnisse genau sind
Our results confirm that the stronger the ITP strategy, i.e., the more frequently pruning is applied, the higher is the potential for reducing the number of nonzero weights while maintaining a high prediction accuracy. This is clearly visible in Fig.~\ref{fig:nonzeros:batchw} and Fig.~\ref{fig:nonzeros:epochw}. It shows for representative training runs and for different values of the regularization hyperparameter $\lambda$, how the number of nonzero weights ($\Omega_{l_0}$) decreases over the epochs for batchwise and epochwise pruning, respectively. It can be recognized that batchwise pruning leads to a faster and stronger reduction of the number of nonzero weights. 
However, this difference decreases with an increasing value of the regularization hyperparameter $\lambda$. Furthermore, it can be observed that the higher the value of $\lambda$, the earlier the number of nonzero weights converges, see Fig.~\ref{fig:nonzeros:batchw}.

In Fig.~\ref{fig:violin} the distribution of weight values over a training process is shown via violin plots. The eight histograms represent the distribution of weights in \texttt{layer3}: %TODO: Warum in (a) und (b) layer1 und in (c) layer3?
how they have been initialized, before the training, after every fifth epoch and after the training (with a regularization hyperparameter of $\lambda=0.00005$). Note that the \(l_1\)-regularization causes the sum of the absolute values of the weights to decrease, and the pruning strategy leads to a large number of weights that are exactly zero. Fig.~\ref{fig:violin} clearly shows that a large majority of weights is relatively close to zero. Actually there are no weights with values in the interval $[-\tau,\tau]$ (here, $\tau=0.001$), which can't be seen in Fig.~\ref{fig:violin} due to the scaling. 

Both Fig.~\ref{fig:CEversusL1(1.2)} and Fig.~\ref{fig:L0versusAcc(1.2)} show the trade-off between the respective loss and regularization terms. The results indicate that the number of nonzero weights (\(\Omega_{l_0}\)), respectively their sum of absolute values (\(\Omega_{l_1}\)),  can be considerably reduced without significantly deteriorating prediction accuracy and cross entropy.  
Indeed, the prediction accuracy can be maintained at a very high level of more than $98$\% even if only about $1,250$ out of the approximately $48,000$ considered \texttt{layer1}-weights are nonzero (about $2.6$\%).
Note that similar outcome vectors are obtained for a range of $\lambda$-values. For ITP training, for example, excellent results are obtained for regularization hyperparameter values $\lambda\in[0.0001,0.0007]$, see, for example,  Tab.~\ref{tab:batchw:mnist}. 

\begin{table}[tbhp]
\caption{The impact of the regularization hyperparameter $\lambda$ and different pruning strategies.\label{tab:prun}}
\scalebox{0.6}{
\footnotesize\noindent
\begin{tabular}{l@{\extracolsep{0.5cm}}l}
\subfigure[MNIST on LeNet-$5$ with \emph{after training} pruning and threshold $\tau = 0.001$.\label{tab:afterw:mnist}]{
    \begin{tabular}{|l|SS|S@{\hspace{-2ex}}|s[table-unit-alignment = right]s[table-unit-alignment = right]s[table-unit-alignment = right]|SS|} 
        \hline
        \multicolumn{1}{|c|}{$\lambda$} & \multicolumn{2}{c|}{$E_{\text{CE}}$} & \multicolumn{1}{c|}{$\Omega_{l_1}$} & \multicolumn{3}{c|}{$\Omega_{l_0}$} & \multicolumn{2}{c|}{Acc. \rule{0pt}{2.5ex}} \\
        & \multicolumn{1}{c}{train} & \multicolumn{1}{c|}{test} & &
        \multicolumn{1}{c}{$\hat{w}^1$} & \multicolumn{1}{c}{$\hat{w}^2$} & \multicolumn{1}{c|}{$\hat{w}^3$} &
        \multicolumn{1}{c}{train} & \multicolumn{1}{c|}{test}\\ \hline
        0&0.004891044857620674&0.06885890818049878&10670,6266&47794&10045&837&0.9984833598136902&0.9873999953269958\\\hline
        $1\cdot 10^{-5}$ &0.01032748189094166&0.037393349126204845&2275.1253&18188&4474&616&0.9977333545684814&0.9904000163078308\\
        $3 \cdot 10^{-5}$ &0.014531460263455906&0.03539449798874557&1102.7369&11424&2444&539&0.996999979019165&0.9905999898910522\\
        $5 \cdot 10^{-5}$ &0.029253905055468203&0.04898362473005546&795.1038&8140&1881&476&0.9918000102043152&0.9868999719619751\\
        $7\cdot 10^{-5}$ &0.024192122031897315&0.0408528646247089&629.9103&8171&1592&412&0.9944333434104919&0.9883000254631042\\
        $9\cdot 10^{-5}$ &0.03036064626644365&0.04452291917274706&507.7365&6086&1222&398&0.9926666617393494&0.988099992275238\\\hline
        $1\cdot 10^{-4}$ &0.028353060801823933&0.04931578964591026&499.2816&6356&1337&364&0.9933833479881287&0.9871000051498413\\
        \hl $3 \cdot 10^{-4}$ &\hl 0.04843172872620635 &\hl  0.04908654497223906 & \hl  215.5405 & \color{gray} 3668 &\color{gray} 459 & \color{gray} 203 &  \hl 0.9884666800498962 &\hl  0.9883999824523926\\
        $5\cdot 10^{-4}$ &0.06476497446619905&0.06432856749617494&158.3316&3167&937&341&0.9836500287055969&0.9825000166893005\\
        $7\cdot 10^{-4}$ &  0.07555826431926885 & 0.07313675489044982 & 119.3474 &  2399 &  675  &  266 & 0.9819166660308838 & 0.9822999835014343\\
        $9\cdot 10^{-4}$ &0.09539416419764361&0.09692352101504804&109.4407&2245&308&199&0.9757999777793884&0.9751999974250793\\ \hline
        0.001&0.09522047584979638&0.08874271142610814&106.5666&2226&625&251&0.9771000146865845&0.9771000146865845\\
        0.003&0.16583310644427934&0.14988349972367287&46.7591&893&179&150&0.9572666883468628&0.9617999792098999\\
        0.005&2.301192618942261&2.301097928237915&0.0&0&0&0&0.11236666887998581&0.11349999904632568\\
        0.007&2.3011927413309925&2.3010632842387073&0.0&0&0&3&0.11236666887998581&0.11349999904632568\\
        0.009&2.301248368962606&2.301167012023926&0.0&0&0&0&0.11236666887998581&0.11349999904632568\\
        \hline
    \end{tabular}
 }
 &
 \subfigure[MNIST on LeNet-$5$ with \emph{epochwise} pruning and threshold $\tau = 0.001$.\label{tab:epochw:mnist}]{
    \begin{tabular}{|l|SS|S@{\hspace{-2ex}}|s[table-unit-alignment = right]s[table-unit-alignment = right]s[table-unit-alignment = right]|SS|} 
        \hline
        \multicolumn{1}{|c|}{$\lambda$} & \multicolumn{2}{c|}{$E_{\text{CE}}$} & \multicolumn{1}{c|}{$\Omega_{l_1}$} & \multicolumn{3}{c|}{$\Omega_{l_0}$} & \multicolumn{2}{c|}{Acc.\rule{0pt}{2.5ex}} \\
        & \multicolumn{1}{c}{train} & \multicolumn{1}{c|}{test} & &
        \multicolumn{1}{c}{$\hat{w}^1$} & \multicolumn{1}{c}{$\hat{w}^2$} & \multicolumn{1}{c|}{$\hat{w}^3$} &
        \multicolumn{1}{c}{train} & \multicolumn{1}{c|}{test}\\ \hline
        0&0.0036283321158572476&0.04823019504712637&10780.1845&47817&10050&837&0.9985833168029785&0.9896000027656555\\ \hline
        $1 \cdot 10^{-5}$&0.010998121394415698&0.04535967440873384&2540.0298&22114&5596&669&0.9975333213806152&0.988099992275238\\
        $3 \cdot 10^{-5}$ & 0.01700537635862827 &0.04035876208722591&1173.8608&12793&2801&534&0.996399998664856&0.9887999892234802\\
        $5 \cdot 10^{-5}$  & 0.04405863972902298 &0.06005888652801514&809.7639&9158&1908&409&0.9867333173751831&0.9824000000953674\\
        $7 \cdot 10^{-5}$ & 0.024235164674186772&0.037224180672629746&632.2781&8293&1470&383&0.9947999715805054&0.9886999726295471\\
        $9 \cdot 10^{-5}$ &0.026849076554775232&0.04155810892343521&518.2909&6706&1130&366&0.9940166473388672&0.9890999794006348\\ \hline
        $1 \cdot 10^{-4}$ &0.028191229472060993&0.039485053570568564&452.2224&5796&1013&366&0.9940500259399414&0.9890000224113464\\
        \hl $3 \cdot 10^{-4}$  &\hl 0.052690875892341144 &\hl  0.05882826953381301 &\hl  210.2285 &\color{gray} 2975 & \color{gray}  443 & \color{gray}  307 &\hl  0.9872666597366333 &\hl 0.9846000075340271\\
        $5 \cdot 10^{-4}$ &0.06767626471779076&0.06601052513382163&151.0716&2959&674&273&0.9837499856948853&0.9828000068664551\\
        $7 \cdot 10^{-4}$  & 0.07597380040129648 & 0.0766559845729731 & 124.0723 &  2562 & 667 & 239 & 0.9817000031471252 &  0.9790999889373779\\
        $9 \cdot 10^{-4}$ &0.10299298301610009&0.09617376556468664&104.1317&2118&401&172&0.9734166860580444&0.9746999740600586\\ \hline
        0.001&0.08794213665693533&0.08436227468221914&99.4126&1726&371&251&0.9783666729927063&0.978600025177002\\
        0.003&0.1578367344104452&0.14708871357075404&48.0325&916&485&150&0.9603000283241272&0.9617999792098999\\
        0.005&0.17535279512802762&0.165300502717495&32.2764&602&150&97&0.9566166400909424&0.9574999809265137\\
        0.007&2.301202852270504&2.301100179947913&0.0&0&0&5&0.11236666887998581&0.11349999904632568\\
        0.009&2.3012099989310526&2.3011118446087466&0.0&0&0&5&0.11236666887998581&0.11349999904632568\\
        \hline
    \end{tabular}
  }
  \\\\
  \subfigure[MNIST on LeNet-$5$ with \emph{batchwise} pruning and threshold $\tau = 0.001$. \label{tab:batchw:mnist}]{
    \begin{tabular}{|l|SS|S@{\hspace{-2ex}}|s[table-unit-alignment = right]s[table-unit-alignment = right]s[table-unit-alignment = right]|SS|} 
        \hline
        \multicolumn{1}{|c|}{$\lambda$} & \multicolumn{2}{c|}{$E_{\text{CE}}$} & \multicolumn{1}{c|}{$\Omega_{l_1}$} & \multicolumn{3}{c|}{$\Omega_{l_0}$}  & \multicolumn{2}{c|}{Acc.\rule{0pt}{2.5ex}} \\
        & \multicolumn{1}{c}{train} & \multicolumn{1}{c|}{test} & &
        \multicolumn{1}{c}{$\hat{w}^1$} & \multicolumn{1}{c}{$\hat{w}^2$} & \multicolumn{1}{c|}{$\hat{w}^3$} &
        \multicolumn{1}{c}{train} & \multicolumn{1}{c|}{test}\\ \hline
        0&0.006185093239570658&0.045581444384530184&10542.293&46551&\;9830&\, 830&0.9989333152770996&0.9904999732971191\\ \hline
        $1 \cdot 10^{-5}$&0.00947833003739516&0.03686343873679637&2483.2019&17751&4575&614&0.9980833530426025&0.9904999732971191\\
        $3 \cdot 10^{-5}$&0.016941222798700133&0.0379129890460521&1227.3847&12582&2434&503&0.996566653251648&0.9878000020980835\\
        $5 \cdot 10^{-5}$&0.022610674143830937&0.04179233047664165&833.9528&9260&1689&414&0.9949833154678345&0.9891999959945679\\
        $7 \cdot 10^{-5}$&0.030985005473792557&0.04820279973715544&610.6045&5946&1192&365&0.9918333292007446&0.9872000217437744\\
        $9 \cdot 10^{-5}$&0.02561723039922615&0.043718430971875784&523.6440&5779&977&298&0.994533360004425&0.9894000291824341\\\hline
        $1 \cdot 10^{-4}$ &0.02772975876828035&0.04608256263434886&500.3722&5491&907&288&0.9937000274658203&0.988099992275238\\
        \hl  $3 \cdot 10^{-4}$ & \hl 0.04919724216957888 & \hl  0.05691415538489819 &  \hl 208.3993 & \color{gray} 2107 & \color{gray}  300 &\color{gray}  170 & \hl 0.989133358001709 & \hl 0.9847999811172485\\
        $5 \cdot 10^{-4}$&0.06400131442844867&0.06507672249376775&143.6565&1408&179&117&0.9849500060081482&0.9829999804496765\\
        $7 \cdot 10^{-4}$&    0.07739979026913643 &   0.07789476969838141 &   115.6429 & 1253 & 157 & 116 &   0.9817166924476624 &  0.9814000129699707\\
        $9 \cdot 10^{-4}$&0.08552288647790748&0.08148979120552541&101.7949&1176&178&95&0.9791333079338074&0.980400025844574\\\hline
        0.001&0.09140587946871917&0.0835100808531046&86.3949&877&99&80&0.9784333109855652&0.9793999791145325\\
        0.003&0.13438978502949075&0.12446605107188222&44.5467&359&62&57&0.9712499976158142&0.9721999764442444\\
        0.005&2.301501661300659&2.30146128616333&0.0&0&0&0&0.11236666887998581&0.11349999904632568\\
        0.007&2.301486335500081&2.3014917587280275&0.0&0&0&0&0.11236666887998581&0.11349999904632568\\
        0.009&2.30125551516215&2.3009899097442625&0.0&0&0&0&0.11236666887998581&0.11349999904632568\\
        \hline
    \end{tabular}
  }
  &
  \subfigure[CIFAR$10$ on VGG with \emph{batchwise} pruning and threshold $\tau = 0.001$.\label{tab:batchw:cifar}]{
  \begin{tabular}{|l|SS|S@{\hspace{-2ex}}|s[table-unit-alignment = right]|SS|}\hline
        \multicolumn{1}{|c|}{$\lambda$} & \multicolumn{2}{c|}{$E_{\text{CE}}$} & \multicolumn{1}{c|}{$\Omega_{l_1}$}&  \multicolumn{1}{c|}{$\Omega_{l_0}(w^3)$} & \multicolumn{2}{c|}{Acc.\rule{0pt}{2.5ex}} \\
        & \multicolumn{1}{c}{train} & \multicolumn{1}{c|}{test} & & &
        \multicolumn{1}{c}{train} & \multicolumn{1}{c|}{test}\\ \hline
        0&0.2695674688577652&0.4295621687889099&12402.742065429688& 20199&0.9416000247001648&0.8942999839782715\\  \hline
            $1 \cdot 10^{-4}$&0.30246205734252934&0.448085977935791&375.081600189209&3972&0.9347599744796753&0.8870999813079834\\
        %0.0002&0.32390345438003537&0.4670829262733459&167.27833366394043&2176&0.9281399846076965&0.8841999769210815\\
            $3 \cdot 10^{-4}$&0.3104508005809784&0.4506808609962463&115.19174194335938&1584&0.9321799874305725&0.8894000053405762\\
        %0.0004&0.3339308268260956&0.47597453832626346&85.58672332763672&1274&0.9236400127410889&0.8802000284194946\\
            $5 \cdot 10^{-4}$&0.32553162330627444&0.4603520473480225&69.53332281112671&1100&0.9271000027656555&0.8859999775886536\\
        %0.0006&0.3168027952861786&0.46254544682502746&59.04883861541748&993&0.9298400282859802&0.8880000114440918\\
            $7 \cdot 10^{-4}$&0.3178138580513001&0.4555770296573639&52.04949903488159&885&0.9285799860954285&0.8870999813079834\\
        %0.0008&0.3171383656787872&0.45747496166229246&45.6668496131897&798&0.9306399822235107&0.8914999961853027\\
            $9 \cdot 10^{-4}$&0.3224200718450546&0.4551865603208542&41.585649251937866&744&0.9285399913787842&0.8881000280380249\\\hline
        0.001&0.3441664080810547&0.47878327355384825&37.17301678657532&659&0.9233999848365784&0.8819000124931335\\
        %0.002&0.3257270219135285&0.4607348445415497&21.34886395931244&487&0.9287999868392944&0.887499988079071\\
        0.003&0.3125132464933395&0.4552609966516495&15.906402468681335&443&0.9326000213623047&0.8883000016212463\\
        %0.004&0.3260964339065552&0.46715286889076235&12.516441345214844&386&0.9277600049972534&0.8859999775886536\\
        0.005&0.3262600348591804&0.4614822541356087&10.736970841884613&376&0.9294599890708923&0.8873000144958496\\
        %0.006&0.31689040689945225&0.44813681204319&9.264301598072052&363&0.930180013179779&0.8927000164985657\\
        0.007&0.3397882418322563&0.4770789935469627&8.445173680782318&359&0.9225999712944031&0.8823000192642212\\
        %0.008&0.3261489935779572&0.4634078051090241&7.665024757385254&335&0.9282799959182739&0.8906999826431274\\
        0.009&0.3337183974051475&0.4701347704291344&6.971011221408844&312&0.9254000186920166&0.8878999948501587\\\hline
        0.01&0.3308692530155182&0.4726362589359283&6.478066325187683&305&0.9251999855041504&0.8830000162124634\\
        %0.02&0.34780799067497253&0.47289502463340755&3.8100494742393494&246&0.9192600250244141&0.8848000168800354\\
        \hl 0.03 & \hl  0.36049116160154343 & \hl  0.4913277043223381 & \hl 2.7803162038326263 & \color{gray} 209 & \hl 0.9174000024795532 & \hl 0.8795999884605408\\
        %0.04&0.36513570297241205&0.49564732227325436&2.1813998222351074&181&0.9149399995803833&0.8805000185966492\\
        0.05&0.3938525557267666&0.5301759891450405&1.9016118794679642&171&0.909060001373291&0.8704000115394592\\
        %0.06&0.36435073182582856&0.4981512862443924&1.6100550591945648&159&0.9162799715995789&0.8783000111579895\\
        0.07&0.3820543394708633&0.5161518308877945&1.4531930088996887&151&0.9097999930381775&0.8712999820709229\\
        %0.08&0.3759251289844513&0.5005263891696929&1.288186103105545&142&0.9122400283813477&0.8773999810218811\\
        0.09 &  0.3995957257127762 &  0.5264757225275041 & 1.1709217429161072 & 137 & 0.9055799841880798 &  0.8701000213623047\\
        \hline
  \end{tabular}\hspace*{1cm}
  }
  \end{tabular}
}
\end{table} 

The impact of the regularization hyperparameter $\lambda$ on the effectiveness of the pruning strategy and on the prediction accuracy is summarized in Tab.~\ref{tab:afterw:mnist} -- \subref{tab:batchw:mnist}. The rows show exemplary results of independently performed training runs for different values of the regularization hyperparameter $\lambda$. 
We observe that the prediction accuracy deteriorates only slightly up to a value of $\lambda=0.0003$  while the number of nonzero weights decreases significantly. For this \emph{critical} value of the regularization hyperparameter, $\lambda^* = 0.0003$, the total number of nonzero weights in all dense layers was reduced in batchwise pruning by about $96$\% and in epochwise and after training pruning only by about $93$\%. In all tables the row highlighted in gray corresponds to a critical $\lambda^*$ beyond which the prediction accuracy deteriorates significantly. An efficient identification of this critical value and of the associated knee solution on the Pareto front is thus highly desirable.

Note that when comparing the three dense layers of LeNet-$5$ pruning is more effective on layers with a larger number of weights.
For example, in epochwise pruning the dense \texttt{layer3} was reduced by only about $64$\% while the larger dense \texttt{layer1} and \texttt{layer2} were reduced by about $94$\% to $96$\%, respectively. 

\subsubsection{Test Results for ITP on VGG}\label{subsubsec:VGG}
%%1.was gemacht wurde
Additional tests for the CIFAR$10$ dataset with a VGG-like network architecture confirm the observations reported in  Section~\ref{subsubsec:LeNet} on the positive effects of ITP. In the case of batchwise pruning the complexity reductions are even more significant than for MNIST, see Tab.~\ref{tab:batchw:cifar}. Indeed, a reduction of almost $99$\% of nonzero weights is achieved in the dense layer of the CNN model while we lose less than $2$\% in the prediction accuracy on the test data.

\begin{figure*}[tbhp]
\subfigure[Model loss for  $\lambda=0.03$ using LRS.  \label{fig:loss}]{%
  \includegraphics[width=0.32\textwidth,trim=0 0 35 0,clip]{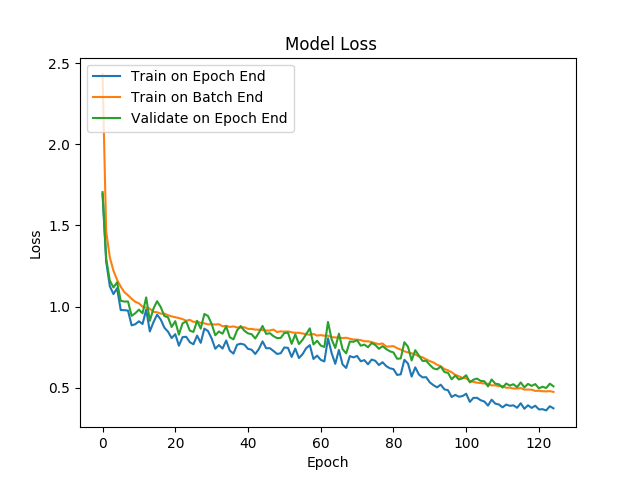}}
  \hfill
\subfigure[Prediction accuracy for $\lambda=0.03$ using LRS.\label{fig:acc}]{%
  \includegraphics[width=0.32\textwidth,trim=0 0 35 0,clip]{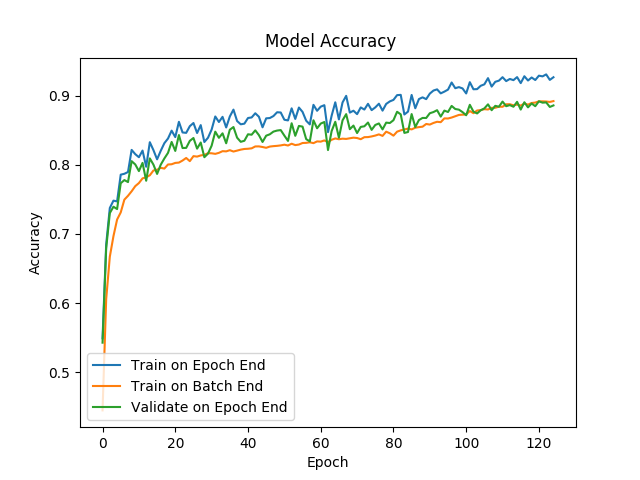}}
\hfill
\subfigure[$\Omega_{l_0}(\hat{w})$ for $\lambda=0.004$, different thresholds $\tau$ and 
LRS.
\label{fig:lrvsnonz}]{
  \includegraphics[width=0.32\textwidth,trim=0 0 35 0,clip]{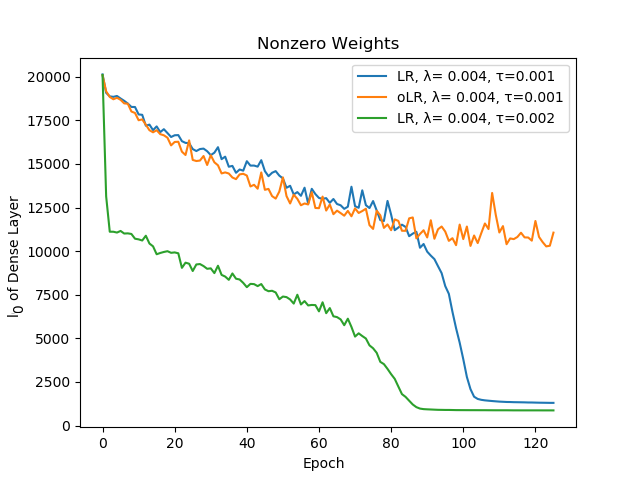}}
\caption{VGG-like network on CIFAR$10$ trained for $\kappa_{\text{max}}=125$ epochs. The pruning threshold for a) and b) is $\tau=0.001$ while in c) two different thresholds are compared with fixed and varying learning rate schedules. }
\end{figure*}

Epochwise and after training pruning are again less effective. A critical value for the regularization hyperparameter is $\lambda^*=0.03$, for which a reduction of nonzero weights in the \texttt{denselayer} by about $98.9$\% is realized. Since on CIFAR$10$ it is much more difficult to achieve good prediction accuracies, we additionally initialize the training process (i.e., the RMSProp optimizer) with a decreasing learning rate which decreases from $t_{\text{start}}=0.001$ to $t_{\text{end}}=~0.0001$ as shown in Fig.~\ref{fig:learningrate}. 
%%2.was die Ergebnisse genau sind
This induces a second steep decrease of the loss value and a corresponding increase of the prediction accuracy as can be seen  Fig.~\ref{fig:acc} and Fig.~\ref{fig:loss}, respectively.
It not only leads to a more accurate performance in training and validation data, but also to a higher pruning effect which can be seen in Fig.~\ref{fig:lrvsnonz}. 
We compare two different pruning thresholds $\tau=0.001$ and $\tau=0.002$. The results shown in Fig.~\ref{fig:lrvsnonz} suggest that the impact of the LRS is also influenced by the threshold value.

One can see clearly that with the same $\lambda$ but different thresholds ($\tau=0.001$ and $\tau=0.002$) the epoch from which on the majority of weights falls below the threshold differs significantly. Indeed, here the LRS has a strong and favorable effect. Moreover, larger threshold values induce a faster convergence towards sparse networks. 

\subsection{Multiobjective Training} \label{subsec:exp_multiobj}
The following experiments all integrate batchwise ITP to reinforce the minimization of the second regularizing objective function $\Omega_{l_1}$. While different variants of the SMGD algorithm are tested both on MNIST and CIFAR$10$  datasets, see Section~\ref{subsubsec:exp_SMGD}, we restrict ourselves to LeNet-$5$ training on MNIST when approximating knee solutions in Section~\ref{subsubsec:exp_knee} by iteratively solving scalarized subproblems with the SGD algorithm.

\subsubsection{Stochastic Multi-Gradient Descent}\label{subsubsec:exp_SMGD}
As our implementation of SMGD is built into the widely used open source SGD methods of Keras (including the Adam and RMSProp extensions) it inherits the same features as their vanilla implementations. Added and new features of our implementation are the consideration of the regularization term as a second objective function and a corresponding built-in update scheme for the weighting parameters according to Algorithm~\ref{alg:smgd}, such that in each iteration a direction of multiobjective steepest descent is chosen. 
We compare different features like the use of a momentum, decay and/or a learning rate schedule. 

\begin{figure*}[htb]
\subfigure[SMGD vs.\ SGD with \newline using $t_{\text{start}}=0.01$ and LRS.\label{fig:acc-smgd+d+m+lr}]
{ 
\includegraphics[width=0.32\textwidth,trim=8 0 40 0,clip]{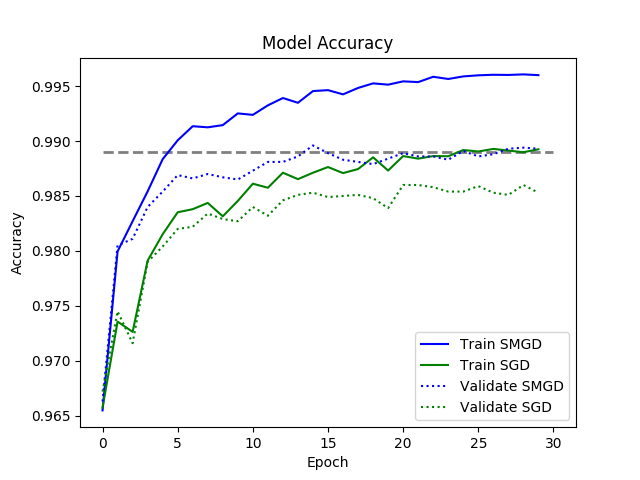}}
  \hfill
\subfigure[MAdam vs.\ Adam with \newline using $t_{\text{start}}=0.001$ and LRS.\label{fig:acc-madam+lr}]{
\includegraphics[width=0.32\textwidth,trim=8 0 40 0,clip]{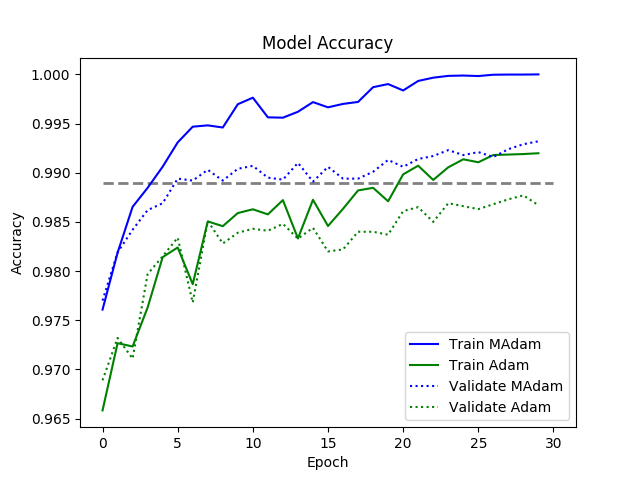}}
\hfill
\subfigure[MRMSProp vs.\ RMSProp\newline with $t_{\text{start}}=0.0001$  and LRS.\label{fig:acc-mrms+lr}]{\includegraphics[width=0.32\textwidth,trim=8 0 40 0,clip]{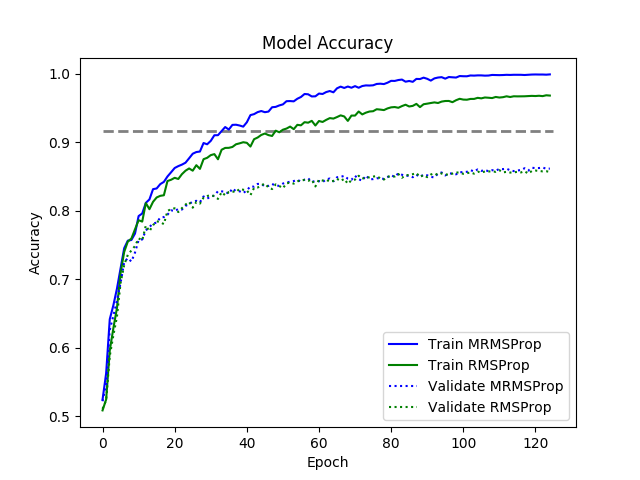}}
\caption{Predicition accuracies for MNIST (Fig.~\ref{fig:acc-smgd+d+m+lr} and Fig.~\ref{fig:acc-madam+lr}) after $\kappa_{\text{max}}=30$ epochs and for CIFAR$10$ (Fig.~\ref{fig:acc-mrms+lr}) after $\kappa_{\text{max}}=125$ epochs. 
\label{fig:acc-comp}}
\end{figure*}

\begin{figure*}[htb]
\subfigure[SMGD vs.\ SGD\label{fig:c2-smgd+d+m+lr}]
{\includegraphics[width=0.32\textwidth,trim=8 0 40 0,clip]
{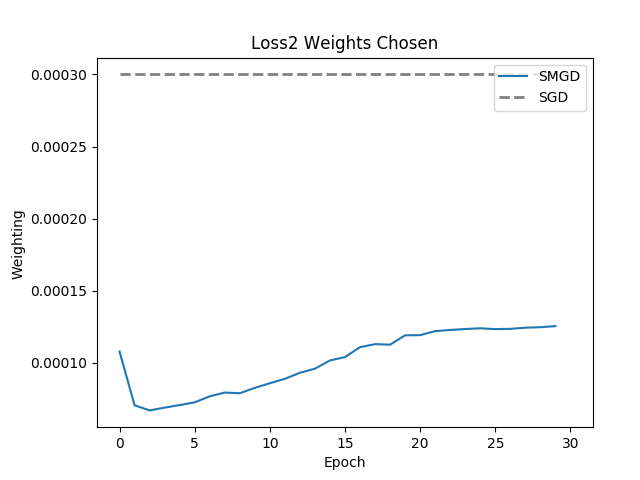}}
  \hfill
\subfigure[MAdam vs.\ Adam\label{fig:c2-madam+lr}]{
 \includegraphics[width=0.32\textwidth,trim=8 0 40 0,clip]
 {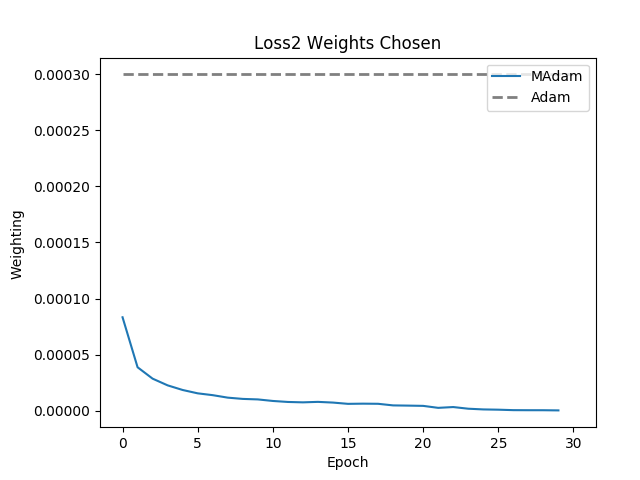}}
\hfill
\subfigure[MRMSProp vs.\ RMSProp\label{fig:c2-mrms+lr}]
{\includegraphics[width=0.32\textwidth,trim=8 0 40 0,clip]
{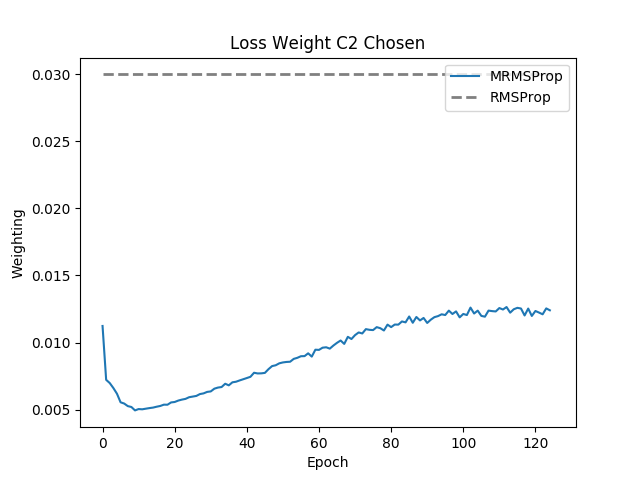}}
\caption{Evolution of the weighting parameter $\lambda$ for the second objective $\Omega_{l_1}$ in SMGD as compared to its fixed value in SGD when training for $\kappa_{\text{max}}=30$ epochs on MNIST (Fig.~\ref{fig:c2-smgd+d+m+lr} and Fig.~\ref{fig:c2-madam+lr}) and for $\kappa_{\text{max}}=125$ epochs on CIFAR$10$ (Fig.~\ref{fig:c2-mrms+lr}).
\label{fig:c2}}
\end{figure*}

The resulting evolution of training- and validation accuracies are shown in Fig.~\ref{fig:acc-comp}. 
For comparison, the dashed line corresponds to the baseline training accuracy that we achieved with a critical weighting parameter $\lambda^*$ as reported in Subsections~\ref{subsubsec:LeNet} and \ref{subsubsec:VGG}, respectively.  
We notice that in all three cases both the training- and the validation accuracy of the respective SMGD algorithms are higher even though, except for the weighting of the objective functions which is constant in the SGD algorithm while it is automatically adapted in the SMGD algorithm, all other settings were the same. This comes, however, at the price of a significantly higher network complexity after SMGD training, i.e., a largely inferior value in the second objective function $\Omega_{l_1}$. Indeed, the weighting parameter $\lambda$ for the second objective function $\Omega_{l_1}$ (c.f.\ \eqref{eq:weightedsum}), that is automatically adapted during SMGD training while it stays fixed in SGD training, remains significantly below the critical value of $\lambda^*$ (that approximates a knee solution) during the complete SMGD training. This is shown in Fig.~\ref{fig:c2}, where  
the dashed line corresponds to the critical weighting parameter $\lambda^*$. 
In general, it is noticeable that the validation accuracy and the training accuracy  are further separated when using SMGD, which is also an indication for overfitting. 

\subsubsection{Approximating the Pareto Knee}\label{subsubsec:exp_knee}
While the SMGD method has proven to converge quickly to an approximate Pareto optimal solution, this solution turns out to be prone to overfitting and usually far from a knee solution. Indeed, the first objective, i.e., the loss function, largely overrides the second regularizing objective function when automatically adapting the weighting parameter in the SMGD method. 

\begin{figure*}[ht]
\subfigure[Pre-defined weighting parameters.\label{fig:paretofront}]
{\includegraphics[width=0.48\textwidth,trim=10 0 40 0,clip]
{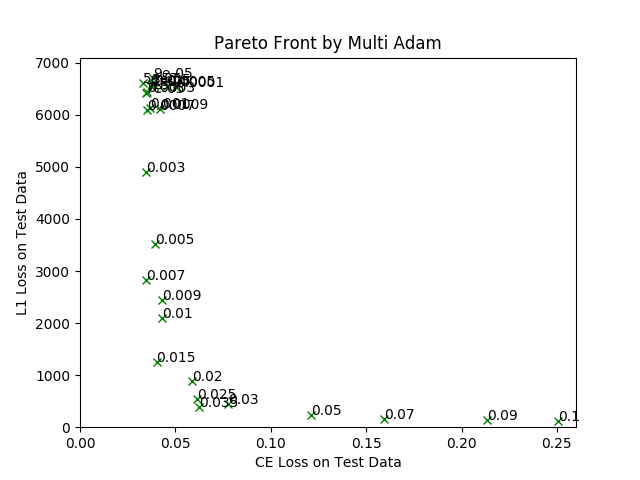}}
\hfill
\subfigure[Dichtomic search. \label{fig:dsheu}]
{\includegraphics[width=0.48\textwidth,trim=10 0 40 0,clip]{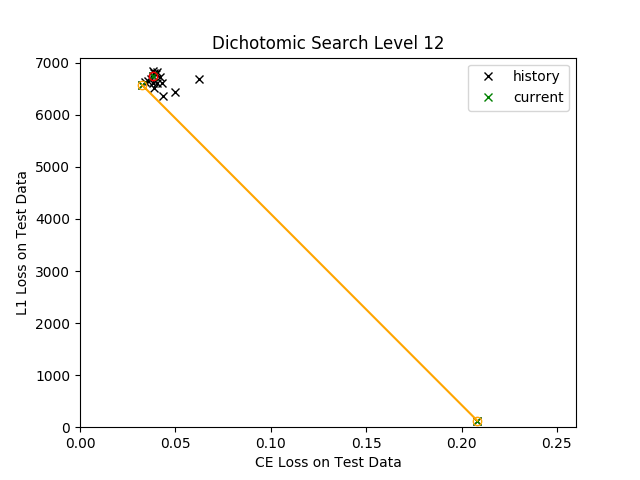}}
\\
\subfigure[Bisection search. \label{fig:bsheu}]
{\includegraphics[width=0.48\textwidth,trim=10 0 40 0,clip]{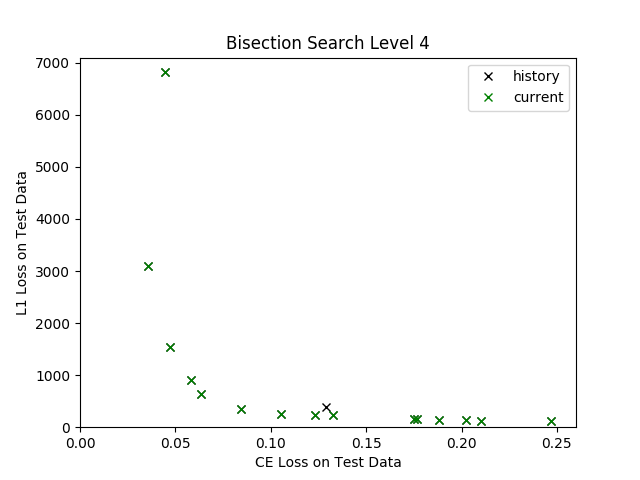}}
\hfill
\subfigure[Stochastic dichotomic search. \label{fig:sdsheu}]
{ \includegraphics[width=0.48\textwidth,trim=10 0 40 0,clip]{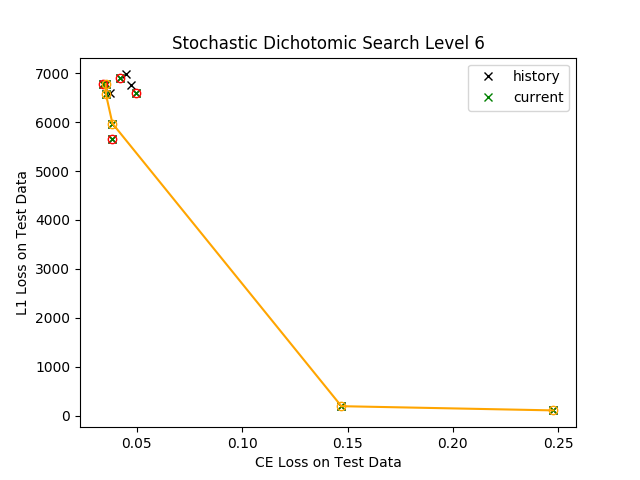}}
\caption{Pareto front approximations using different search heuristics for finding critical weighting parameters and knee solutions.\label{fig:knee}}
\end{figure*}

In biobjective neural network training critical weighting parameters tend to be in a rather small interval close to zero which can be explained by the very different scaling, slope and curvature of the two objective functions. Slightly larger weighting parameters already lead to collapsed networks (with almost all weights equal to zero) while slightly smaller weighting parameters lead to overfitting (with almost no weights equal to zero).
We investigate four different approaches to find critical weighting parameters that are all based on the repeated solution of weighted sum scalarizations \eqref{eq:weightedsum} using a single-objective SGD method. 
The corresponding results are presented in Fig.~\ref{fig:knee}. Except in the first case were significantly more training runs performed, the number of SGD calls and hence the total training cost was nearly the same. 
Moreover, the number of epochs was reduced to save computational time. 
Fig~\ref{fig:paretofront} serves as a benchmark by showing the training results w.r.t.\ a predefined list of weighting parameters from the interval $[0.00001,0.1]$ that are condensed towards smaller values. A pronounced knee is clearly visible. Fig.~\ref{fig:dsheu} and Fig.~\ref{fig:sdsheu} show the results of a classical dichotomic search algorithm and of stochastic dichotomic search according to Algorithm~\ref{alg:dicho}, respectively. While theoretically the classical dichotomic search algorithm should be able to quickly find a knee solution this is not the case in our experiments. This can be explained by the rather small interval of relevant weighting parameters, the large sensitivity w.r.t.\ small variations of the weighting parameter and, even more importantly, by the fact that the SGD is not guaranteed to converge to a Pareto optimal point. The stochastic dichotomic search algorithms shows a slightly better performance. However, it is still outperformed by a bisection search on the weighting parameter as shown in Fig.~\ref{fig:bsheu}.

\section{Conclusion and Outlook}\label{sec:conc}
In this work we take a multiobjective perspective on neural network training and present a straightforward and easy to implement method for unstructured pruning of dense layers in DNNs which is completely integrated in the training process. With this intra-training pruning (ITP) approach, we reduce the number of nonzero weights without a significant effect on the accuracy by adaptively condensing the network to relevant connections. We obtain the most promising results with \emph{batchwise pruning} where weights are pruned after each batch during the training with respect to a predefined threshold value. We use $l_1$-regularization to guide the pruning process and to simultaneously avoid overfitting. 
By re-interpreting regularization as a second and independent training goal, the trade-off between prediction accuracy and network complexity is analyzed. Optimization algorithms using the stochastic multi-gradient descent algorithm and several variants of scalarization based methods are implemented that address 
the specific challenges of neural network training. 

The main advantage of this new approach can be seen in the fact that the network architecture and the network quality are optimized consistently and simultaneously. Moreover, the role of the regularization hyperparameter is thoroughly analyzed which facilitates informed choices of preferable trade-offs.

Different combinations of the learning rate $t$, the pruning threshold $\tau$ and the regularization hyperparameter $\lambda$ are evaluated for two well-established datasets and network architectures for image classification. While batchwise pruning achieved excellent results on MNIST without further fine-tuning, we suggest to combine batchwise pruning with a learning rate schedule for CIFAR$10$. 
The presented techniques result in higher compression rates while maintaining a competitive prediction accuracy, both for MNIST and for CIFAR10.  

The fine-tuning of hyperparameters is a challenging and time-consuming task when designing and training neural networks. While we focus on the regularization hyperparmeter in this work, we plan to extend our studies to include further hyperparameters in an automated machine learning (AutoML) strategy. This also involves a comparison of threshold controlled pruning methods with pruning strategies that aim at a predefined percentage of weights being zero, see, e.g.,  \citet{Zhu2017}. 

Moreover, several other norms and quality measures may be tested  as the second regularizing objective function in bi- or multiobjective neural network training and in combination with ITP. Particularly, $l_p$-regularization terms with $p<1$ seem to be promising to approximate the \(l_0\)-regularizer. However, this leads to nonconvex optimization problems which makes the training problems more complex.

\paragraph{Acknowledgments}
This work has been partially supported by EFRE (European fund for regional development) project EFRE-0400216.

\bibliographystyle{elsarticle-harv}
\bibliography{lit-bit.bib}
\end{document}